\documentclass{iopjournal_custom}

\usepackage{lmodern}
\usepackage[T1]{fontenc}
\usepackage[protrusion=true,expansion=false]{microtype}
\usepackage{hyperref}       
\usepackage{url}            
\usepackage{booktabs}       
\usepackage{amsfonts}       
\usepackage{nicefrac}       
\usepackage{xcolor}         
\usepackage{array}
\usepackage{amsmath}
\usepackage[ruled,vlined,linesnumbered]{algorithm2e}
\usepackage{graphicx}
\graphicspath{{./images/}}
\usepackage{booktabs}
\usepackage{adjustbox}
\usepackage{subcaption}

\usepackage[nameinlink,capitalize]{cleveref}

\usepackage[numbers]{natbib}  

\newcommand{\be}{\begin{equation}}
\newcommand{\ee}{\end{equation}}
\newcommand{\bea}{\begin{eqnarray}}
\newcommand{\eea}{\end{eqnarray}}

\newcommand{\rup}[1]{\left[#1\right]}

\usepackage[normalem]{ulem} 

\newcolumntype{R}[2]{%
	>{\adjustbox{angle=#1,lap=\width-(#2)}\bgroup}%
	l%
	<{\egroup}%
}
\newcommand*\rot{\multicolumn{1}{R{45}{1em}}}

\begin{document}


\title{Bounded Graph Clustering with Graph Neural Networks}

\author{Kibidi Neocosmos$^{1,*}$, Diego Baptista$^2$ and Nicole Ludwig$^1$}

\affil{$^1$University of T\"ubingen,  T\"ubingen, Germany}

\affil{$^2$Graz University of Technology, Graz, Austria}

\affil{$^*$Author to whom any correspondence should be addressed.}

\email{kibidi.neocosmos@uni-tuebingen.de}


\begin{abstract}
	In community detection, many methods require the user to specify the number of clusters in advance since an exhaustive search over all possible values is computationally infeasible. While some classical algorithms can infer this number directly from the data, this is typically not the case for graph neural networks (GNNs): even when a desired number of clusters is specified, standard GNN-based methods often fail to return the exact number due to the way they are designed. In this work, we address this limitation by introducing a flexible and principled way to control the number of communities discovered by GNNs. Rather than assuming the true number of clusters is known, we propose a framework that allows the user to specify a plausible range and enforce these bounds during training. However, if the user wants an exact number of clusters, it may also be specified and reliably returned.
\end{abstract}

\section{Introduction}\label{sec:introduction}

Graph neural networks (GNNs) have become an increasingly popular choice for graph-based tasks (see \cite{stokes2020deep,derrow2021eta,ying2018graph,mirhoseini2021graph} for example applications). Network classification is a common task and requires a model to match a network to its label. To do so, it transforms the input network into an initial guess of the label and then improves that guess based on its similarity to the true label. The transformation requires a component in the graph neural network known as a pooling layer. The purpose of the pooling layer is to reduce the size of the input data so that it may ultimately become a label. We refer the reader to \cite{grattarola_understanding_2022} for a more comprehensive discussion of pooling in graph neural networks.

Recently, community detection has been proposed as a method for pooling (i.e., to reduce the size of the network data) \cite{bianchi_spectral_2020, tsitsulin_graph_nodate}.
Community detection is the process of identifying meaningful collections of nodes (i.e., communities or clusters) in a network so that they can be replaced or summarized by a singular (super)node. Connections between supernodes are then formed from the connections between the communities that they represent. To perform community detection in the pooling layer of a graph neural network, another graph neural network is used. While other classic non-neural network styled methods, in principle, may also be used, this would no longer allow end-to-end learning -- a hallmark of the success of neural networks.

Community detection using graph neural networks was originally developed in the context of a pooling layer to reduce the size of the network. This process is a supervised learning task that requires labeled data. However, community detection is traditionally an unsupervised learning task; there are no labels for the data. When evaluated in a classical community detection setting, graph neural networks present a surprising flaw: they do not return the number of communities specified by the user. \cite{bianchi_spectral_2020} and \cite{tsitsulin_graph_nodate} hint at this problem by their inclusion of regularization terms in the graph neural network's loss function to influence the number of communities returned. The first part of our contribution addresses this problem by enabling a graph neural network to output the specified number of communities.

The need to specify the number of communities when using a community detection algorithm is a common feature \cite{fortunato_community_2016}. As stated previously, community detection is the process of finding meaningful collections of nodes in a network. The task requires both identifying a meaningful collection of nodes and determining a meaningful number of collections. The second problem influences the first in that part of what makes a collection of nodes meaningful is the \emph{number} of collections. To circumvent the difficulty of addressing both problems, many algorithms require that the number of communities be specified and, as such, they focus on finding a meaningful arrangement of nodes within those communities. However, other algorithms are able to address both problems. The second part of our contribution aims to push graph neural networks toward this class of algorithms by removing the need to specify the exact number of communities.

For our contribution, we propose a constraint that we incorporate into the loss function of a graph neural network. The constraint allows the output number of communities to either be specified exactly or within a given range. Our approach extends the capabilities of graph neural networks for community detection by introducing required functionality and additional flexibility.

\section{Related work}\label{sec:related_work}

Our general goal is to augment the loss function of graph neural networks used for community detection. To this end, we build upon notable examples of such graph neural networks, namely: DiffPool \citep{ying_hierarchical_2018}, MinCutPool \citep{bianchi_spectral_2020}, and DMoN \citep{tsitsulin_graph_nodate}.

An earlier work, DiffPool \citep{ying_hierarchical_2018} introduced graph neural networks as a pooling technique within a larger graph neural network used for network classification. It supplemented the loss function of the graph neural network used for pooling with a link prediction function. MinCutPool \citep{bianchi_spectral_2020} built on this idea and introduced the minimum cut problem as a differentiable loss function, learning soft cluster assignments that minimize edges between clusters while maintaining balanced partitions. This approach enables effective graph pooling by incorporating spectral clustering principles directly into the neural network architecture. More recently, DMoN \citep{tsitsulin_graph_nodate} developed a graph neural network approach for community detection that directly optimizes modularity — a measure quantifying the difference between the number of edges within communities and the expected number under a random null model. In contrast to MinCutPool, DMoN was developed as a standalone community detection technique.

We use a similar approach to MinCutPool and, in particular, to DMoN with the addition of a constraint term to the loss function. 

\section{Notation}\label{subset:notation}

Let a graph $G$ be defined by a set of nodes $V$ and edges $E$, where $n$ is the total number of nodes and $m$ is the total number of edges. The adjacency matrix is represented as $\mathbf{A}\in\{0,1\}^{n\times n}$ where $\mathbf{A}_{ij}=1$ indicates an edge between nodes $i$ and $j$ (and 0 indicates no edge). Additionally, let $d_i:=\sum_{i}^n\bold{A}_{ij}$ be the degree of a node $i$ -- the total number of edges starting at node $i$. We consider the cluster assignment matrix $\mathbf{S}\in [0,1]^{n\times c}$ where $c$ is the maximum number of clusters. The cluster assignment matrix indicates which cluster (indexed by the columns) each node (indexed by the rows) is assigned to. Let $C_k$ be the set of nodes in cluster $k\in\{0,1,...,c\}$ and $|C_k|$ is the number of distinct elements in $C_k$. Lastly, let $\mathbf{e}_k$ be the canonical vector of $\mathbb{R}^c$ for some $k \in \{1,...,c\}$.

\section{Community detection in graph neural networks via modularity maximization}\label{sec:cluster_with_gnn}

Similar to neural networks, graph neural networks (GNNs) are non-linear function approximators. However, unlike neural networks, they leverage the topology of the input data via message passing to generate the output. They are commonly used for tasks such as node and graph classification and, although less common, for community detection. We focus on community detection.

In the context of graph neural networks, community detection is formulated as an optimization problem in which the GNN initially assigns nodes to communities and then iteratively improves this assignment according to an objective (or loss) function. The loss function serves as a measure of the quality of the community assignment, encoding the definition of community for which the GNN optimizes. In general, two common quality functions are used: one based on modularity and another based on the minimum cut problem. In this paper, we focus on modularity as the quality function because it is the most widely adopted community measure (in general) and it is used in the current state-of-the-art GNN for community detection \cite{tsitsulin_graph_nodate}.

Modularity \citep{newman_modularity_2006} formulates the idea of a community from a statistical perspective. In general, a community is seen as a densely connected collection of nodes. The question that follows is: what do we mean by "densely connected"? Modularity takes the stance that nodes are densely connected if there are more edges than one would expect to see at random. Mathematically, it is typically represented using $Q$ and formulated as 
\begin{equation}\label{eqn:modularity}
	Q=\frac{1}{2m}\sum_{ij}^n\left[ \bold{A}_{ij} - \bold{P}_{ij} \right] \delta_{ij}(k_i, k_j),
\end{equation}
where $ \bold{P}_{ij}$ is the probability that there is an edge between node $i$ and $j$; $k_i$ indicates that node $i$ is part of cluster $k$. The Kronecker delta function $\delta_{ij}(k_i, k_j)$ is 1 when nodes $i$ and $j$ are part of the same cluster $k$ and 0 when they are part of different clusters. In the classic formulation of modularity, $\bold{P}_{ij}=\frac{d_i d_j}{2m}$ and $\delta_{ij}(k_i, k_j)=\sum_{ij}^{c} s_{ik}s_{jk}$ where $s_{ik}\in\bold{S}$ and $s_{ik}=1$ if node $i$ is assigned to cluster $k$ (0 otherwise). Thus,
\begin{equation}
	Q=\frac{1}{2m}\sum_{ij}^n\left[ \bold{A}_{ij} - \frac{d_i d_j}{2m} \right]\sum_{k}^{c} s_{ik}s_{jk}.
\end{equation}

Note that when all nodes are placed in a single community, the two terms in \eqref{eqn:modularity} cancel each other (since $\sum_{ij} \bold{A}_{ij} = \sum_{ij} \bold{P}_{ij} = 2m$), and thus modularity is zero. Additionally, $\bold{S}\in\{0,1\}^{n\times c}$ has discrete values (either 0 or 1) in the classical formulation. However, a continuous formulation is used ($\bold{S}\in [0,1]^{n\times c}$) in graph neural networks to ensure modularity is differentiable.

An alternative function that measures the quality of community structure is derived from the minimum cut problem. The minimum cut problem is the task of finding a specified number of disjoint subgraphs of a network by removing as few edges as possible. These disjoint subgraphs are seen as the communities of the network because they are densely connected collections of nodes. Here, as opposed to modularity, densely connected simply means more edges inside the community than between communities. Thus, the problem can be reformulated as maximizing the number of edges inside communities. Mathematically, we can express this function as
\begin{equation}\label{eqn:mincut}
	\sum_{k=1}^{c}\frac{\sum_{i,j\in C_k}^n\mathbf{A}_{ij}}{\sum_{i\in C_k,j\notin C_k}^n\mathbf{A}_{ij}},
\end{equation}
where the numerator is the number of edges inside community $k$ and the denominator is the number of edges leaving the same community. Equation~\ref{eqn:mincut} is usually normalized by a constant to prevent the undesirable solution of having a community with $n-1$ nodes and another with $1$ node. When it is normalized by the number of nodes, it is called \emph{RatioCut} \citep{hagen1992new}, and when it is normalized by the edge weights, it is referred to as \emph{Ncut} \citep{shi_normalized_2000}.

Again, we focus on modularity as a measure of community quality.

\subsection{The problem of choosing the number of clusters}\label{subsec:choosing_k}

Many community detection algorithms require the desired number of clusters as an input. Graph neural networks are no different. In practice, however, the specified number of clusters may be interpreted as an upper bound, rather than a guarantee that exactly that many clusters will be found. To understand why this is the case, we need to keep in mind that community detection in graph neural networks is formulated as an optimization problem: the neural network seeks the nodes' cluster assignments that maximize a desired community quality function (for example, modularity). To that end, the entries of the cluster assignment matrix are continuous, not discrete (for the sake of differentiability) and, thus, represent a soft assignment of the nodes. However, we still desire a hard clustering of nodes and therefore need to convert the soft assignment to a hard assignment. An intuitive and common approach to achieve this is to apply the \textit{max} function to the rows of the cluster assignment matrix: we assign each node to the cluster to which it is most likely to belong. Effectively, we treat the continuous values in the cluster assignment matrix as probabilistic memberships, where each entry represents the probability that a node is assigned to a given cluster. And in so doing, we divide nodes into clusters based on the \textit{highest} probability of being assigned to a cluster. Of course, this is not the only way one might transition from soft to hard clustering but, as previously mentioned, it intuitively makes sense, especially considering that our aim in the optimization problem is to find the cluster assignment that maximizes our community quality function. However, when we assign nodes to the cluster they are most likely to belong to, the subsequent number of clusters is at most the number of communities specified ($c$) or fewer. This phenomenon may be exacerbated by a quality function such as modularity, which struggles to identify smaller communities and, as such, is biased towards fewer, larger clusters \cite{fortunato2007resolution, good_performance_2010, fortunato_community_2016}. In our experiments, the graph neural network with only modularity as a loss function consistently finds fewer than the specified number of communities (refer to figures~\ref{fig:varying_cluster_number}, \ref{fig:varying_density}, \ref{fig:varying_network_size}, and \ref{fig:upper_lower})

Given that, in practice, we specify an upper bound on the output number of clusters, one might naturally and prematurely conclude that the graph neural network yields the optimal clustering (and thus the optimal number of clusters) below the upper bound. Ultimately, this is dependent on the quality function used. In the case of modularity, this is rarely the case because it is a non-convex function with multiple near-optimal solutions, making it difficult to optimize \cite{good_performance_2010}.

In light of this problem, researchers typically augment the graph neural network's loss function to influence the number of clusters identified. This is done by adding a regularization term that has two effects: discouraging an undesirable, trivial solution to the quality function and encouraging more balanced clusters (we discuss the regularization terms of the DMoN and MinCutPool models in section~\ref{subsec:dmon_mincut_reg}). The first effect aims to penalize an undesirable solution to the community detection problem, namely, assigning all nodes to a single cluster. It is trivially true that we could maximize the number of intra-community edges if we assign all nodes to a single cluster. This is not a problem if one uses modularity as a quality function because it is, by definition, zero when all nodes are placed in the same cluster. Additionally, the second effect encourages balanced clusters: each cluster should have the same number of nodes, with $c$ clusters in total. Notably, the inclusion of this second effect is not justified in the literature \cite{bianchi_spectral_2020, tsitsulin_graph_nodate}. Perhaps it is to encourage $c$ number of clusters. In our experimentation, it seems to encourage more clusters, although rarely $c$. 

\subsection{Our contribution: bounding the output number of clusters}\label{subsec:solution}

We propose a constraint that enforces a lower bound ($l$) on the number of clusters returned. Inherently, $c$ behaves like an upper bound and, thus, the addition of our constraint creates a range for the output number of clusters. When the upper and lower bounds are equal, the graph neural network will return an exact number of clusters.

The constraint acts on the cluster assignment matrix $\mathbf{S}$. Mathematically, it normalizes each row of $\mathbf{S}$ by the maximum of each row, resulting in $\mathbf{s}'_{ik} = \mathbf{s}_{ik}/\max_{1 \le k \le c}\mathbf{s}_{ik}  $ where $\mathbf{s}_{ik} \in \mathbf{S}$. The $k$-th row entry with value $s'_{ik}=1$ represents the cluster with the highest membership for node $i$. This would also be the cluster to which it belongs if we were to consider only hard membership.
Using this row-normalized matrix, we can define our constraint for the optimization problem as

\begin{align}\label{eqn:constraint}
	\text{constraint}(\mathbf{S},l) &= l - \sum_{k=1}^{l} \max_{1 \le i \le n}\mathbf{s}'_{ik}  \nonumber \\
	&= l - \sum_{k=1}^{l} \max_{1 \le i \le n} \rup{ \mathbf{s}_{ik}/ \max_{1 \le k \le c} \mathbf{s}_{ik}},
\end{align}
where the elements $\max_{1 \le i \le n} \mathbf{s}'_{ik}$ are sorted from largest to smallest and $l$ is the specified lower bound. For further ease of comprehension, we provide an algorithmic description of our constraint in algorithm~\ref{alg:min_num_constraint}.

\begin{algorithm}[ht]
	\caption{Constraint for finding the minimum number of clusters -- constraint$(\bold{S},l)$}
	\label{alg:min_num_constraint}
	\KwIn{The cluster assignment matrix {$\bold{S} $} and the desired lower bound $l$}
	\KwOut{Difference between $l$ and $p\in \mathbb{R}_+$, the predicted number of clusters. Note that $p\leq l$}
	Normalize rows of $\bold{S}$ by the largest element of each row\;
	Find the largest element in each column of the row-normalized $\bold{S}$\;
	$p \gets$ sum the $l$ largest elements\;
	\Return $l-p$\;
	
\end{algorithm}
The constraint serves as a continuous count of the number of communities in relation to a specified lower bound. Intuitively, it may be understood as a continuous measure of how close empty clusters are to becoming non-empty. When finding the largest element in each column, a non-empty cluster is represented as 1, and an empty cluster has a value between 0 and 1, indicating that none of the nodes has its maximum membership in that cluster. The closer the value is to 1, the closer it is to becoming non-empty. In this way, the constraint identifies the empty clusters most likely to become non-empty and encourages them to have at least one member. We also offer an upgraded version of our constraint (equation~\ref{eqn:constraint}) that allows the user to specify the minimum number of nodes per cluster. We include it in the appendix (algorithm~\ref{alg:min_k_min_size_constraint}).

The constraint has computational complexity $\mathcal{O}(n(c + l))$, which is linear in the number of nodes $n$. Specifically, computing normalization factors requires $\mathcal{O}(nc)$ operations, and evaluating the constraint over $l$ clusters requires $\mathcal{O}(ln)$ operations. This linear scaling is small compared to the quadratic complexity $\mathcal{O}(n^2c)$ from the modularity computation (given that $c$ is much smaller than $n$), which dominates the overall computational cost.

In addition to equation~\ref{eqn:constraint}, we introduce a regularization term to encourage balanced clusters (i.e., clusters of equal size). It is a modified version of the MinCutPool\cite{bianchi_spectral_2020} regularization term, which is easier to satisfy since it no longer requires the column vectors of the clusters assignment matrix to be orthogonal. It may be described as
\begin{equation}\label{eqn:balance_reg}
	\text{balance}(\mathbf{S})=\frac{||\text{diag}(\mathbf{S}^T\mathbf{S}) - \frac{n}{c}\mathbf{1}_c||_2}{||n\mathbf{e}_k- \frac{n}{c}\mathbf{1}_c||_2},
\end{equation}
where $\mathbf{1}_c \in \mathbb{R}^c$ denotes the vector of all ones, $\mathbf{e}_k \in \mathbb{R}^c$ is the $k$-th canonical vector, and $\|\cdot\|_2$ denotes the $\ell_2$ norm (Euclidean norm). The denominator normalizes the function to have a maximum value of 1. As such, it attains its maximum value of 1 when all nodes are in a single cluster, and its minimum value of 0 when each cluster has an equal number of nodes.
For our experiments, it is important to include the balanced-cluster regularization with the constraint. The constraint only requires that there is one node per cluster and singleton clusters have a poor modularity. Introducing the balanced-cluster regularization term encourages more nodes per cluster. 

We combine equations~\ref{eqn:modularity}, \ref{eqn:constraint}, and \ref{eqn:balance_reg} into a single loss function that may be used in a graph neural network:
\begin{equation}\label{eqn:loss}
	L(\mathbf{S},\mu,\lambda, l) = - \text{modularity}(\mathbf{S}) + \mu\cdot \text{constraint}(\mathbf{S},l) + \lambda\cdot \text{balance} (\mathbf{S}),
\end{equation}
where the constants $\mu$ and $\lambda$ adjust the influence of the constraint and balance regularization terms, respectively. Note that maximizing modularity is equivalent to minimizing the negative of modularity. As such, modularity is negative in equation \ref{eqn:loss} since it is optimized in graph neural networks using gradient descent.

In summary, we propose a constraint (equation~\ref{eqn:constraint}) that enforces a lower bound $l$ and, coupled with the upper bound $c$, enforces a range for the output number of clusters. The range may be tightened at the user's discretion, with the case $l=c$ yielding an exact number of clusters. We incorporate the constraint into the loss function of a graph neural network used for community detection, along with a regularization term (equation~\ref{eqn:balance_reg}) that encourages balanced clusters. Table~\ref{tab:comparison} summarizes our contribution in the context of prominent GNN architectures upon which our work developed. 

In addition to creating a fixed range, a lower bound may also be used to search for an optimal number of clusters by progressively tightening the range and selecting the output number of clusters with the highest modularity. We do not explore such a procedure in this work and leave it as an open direction of inquiry. Instead, we posit the idea to further highlight the benefit of imposing a constraint on the returned number of clusters.

\begin{table}[h!]
	\caption{A summary of our contribution in relation to different GNN methods for community detection}
	\centering
	\begin{tabular}{lccc}
		\toprule
		\textbf{Method} & \multicolumn{3}{c}{\textbf{Loss}} \\
		\cmidrule(lr){2-4}
		& \textbf{Community Measure} & \textbf{Regularization} & \textbf{Constraint} \\
		\midrule
		MinCutPool & Minimum Cut & $\checkmark$ & $\times$ \\
		\midrule
		DMoN & Modularity & $\checkmark$ & $\times$ \\
		\midrule
		Approach taken in paper & Modularity & $\checkmark$ & $\checkmark$ \\
		\bottomrule
	\end{tabular}
	\label{tab:comparison}
\end{table}

\section{Experiments}

We assess our proposed solution on both synthetic and real-world datasets. 

In our experiments, we compare the performance of four graph neural networks that differ only in their loss functions. The first model uses a loss function based solely on the community quality function modularity; we refer to it as \emph{GNN}. It serves as our baseline and provides a reference point for comparison. The second model is the standard approach currently in use. Its loss function combines a quality function with a regularization term that encourages balanced clusters, and we refer to it as \emph{GNN+REG}. The third model extends \emph{GNN} by incorporating our constraint, and we refer to it as \emph{GNN+CONSTRAINT}. It provides an additional baseline for illustrating only the effect of the constraint. Finally, our proposed method augments the loss function with both the regularization term for balanced clusters, and our constraint that enforces a minimum number of clusters. We refer to this model as \emph{GNN+REG+CONSTRAINT}.  Using models with different terms in the loss function helps delineate the effect of the balance-cluster regularization term and the constraint.

Although each model differs in the loss function, they share a common model architecture. Each model consists of a graph neural network that transforms the input data to a transitional embedding that is then fed to a multilayer perceptron (MLP); this outputs the cluster assignment matrix.  We use GraphSage (with mean aggregation) \citep{hamilton_inductive_2017} as our graph neural network of choice because it is a common and well-established model. Additionally, we set the parameter constants $\mu=\lambda=1$ from equations~\ref{eqn:constraint} and \ref{eqn:balance_reg}. We also use a random weight initialization for both the graph neural network and the multilayer perceptron. As is standard, we initialize each model 3 times to assess the consistency of model performance (i.e., how much of the performance is influenced by the random weight initialization).
\begin{table}[h!]
	\caption{A summary of the synthetic network statistics. 10 networks are generated for each network type and, as such, the number of edges and density are approximations.}
	\centering
	\begin{tabular}{cccccc}
		\toprule
		Size label & Clusters & Density label & Nodes 				& Edges 					& Density \\
		\midrule
		Small 		& 5 	&Low 			& $10^2$ 				& $\sim$ 450		& $\sim$0.04 \\		
		&		&Medium 	& $10^2$ 				& $\sim$ 1 000		& $\sim$0.1 \\		
		
		\midrule
		Medium & 5				&Low 					& $10^3$					& $\sim$ 25 000		& $\sim$0.02  \\
		& 				&Medium				& $10^3$					& $\sim$ 100 000	&$\sim$ 0.1 \\
		& 				&High					& $10^3$					& $\sim$ 150 000	&$\sim$ 0.15 \\
		
		&10 		&Medium					& $10^3$				& $\sim$ 70 000	&$\sim$ 0.07\\
		
		&20 		&Medium					& $10^3$				& $\sim$ 50 000	&$\sim$ 0.05\\
		
		\midrule
		Large 		& 5			&Low 						& $10^4$				& $\sim$1 000 000 & $\sim$0.01  \\

		\bottomrule
	\end{tabular}
	\label{tab:syn_stats}
\end{table}

\subsection{Results on synthetic data}\label{subsec:results_syn}
To generate synthetic data, we use the stochastic block model (SBM), which we parameterize by the community sizes and the edge probabilities within and between communities, denoted by $p_{\text{in}}$ and $p_{\text{out}}$, respectively. We use unequal cluster sizes to provide a more realistic challenge to our models, given our inclusion of the balance-cluster regularization term. The cluster sizes are generated randomly from a uniform distribution (we provide the cluster sizes in table~\ref{tab:cluster_sizes} in the appendix). The controlled setup allows us to systematically evaluate the performance of our method under varying structural properties of the network.
We vary three main properties in the synthetic networks: network size (considered as the number of nodes), the number of ground-truth communities, and the network density (defined in equation~\ref{eqn:density} in the appendix). For ease of discussion, we provide labels for the size and density of the network. The sizes of the network varies between \emph{small} ($10^2$ nodes), \emph{medium} ($10^3$ nodes) and \emph{large} ($10^4$ nodes). Moreover, the density is labeled as \emph{low}, \emph{medium} and \emph{high}, where \emph{low} $<0.05 \leq$ \emph{medium} $\leq 0.1 <$ \emph{high}. Again, these labels are used solely as a means to facilitate communication. We generated 10 networks for each network type (for example, there are 10 \emph{medium}-sized networks with 5 clusters and \emph{low} density). We summarize the combinations of parameters used in table~\ref{tab:syn_stats}.
Additionally, graph neural networks require node features. Therefore, we use the column vector from the adjacency matrix that is associated with a node as its feature vector. This was done to simply provide topological information to the model and not bias it based on node feature similarity.
\begin{figure}[h!]
	\centering
	
	\begin{subfigure}[b]{0.48\textwidth}
		\centering
		\includegraphics[width=\textwidth]{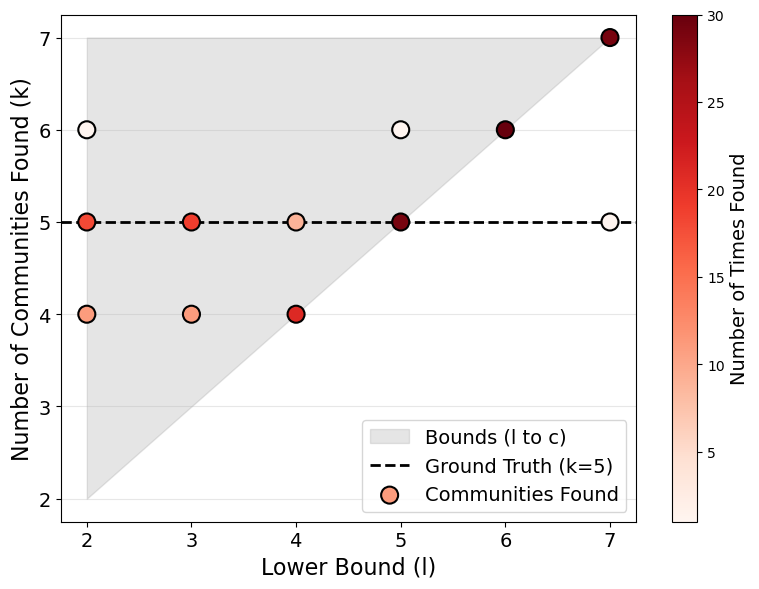}
		\caption{}
		\label{fig:varying_lower_bound_1}
	\end{subfigure}
	\hfill
	\begin{subfigure}[b]{0.48\textwidth}
		\centering
		\includegraphics[width=\textwidth]{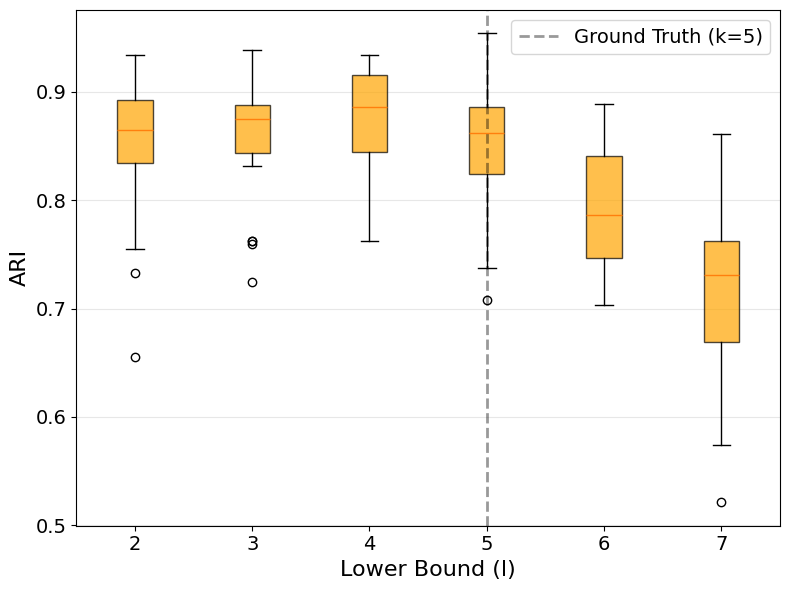}
		\caption{}
		\label{fig:varying_lower_bound_2}
	\end{subfigure}
	
	\caption{(a) The number of communities predicted by model \emph{GNN+REG+CONSTRAINT} as the lower bound varies on \emph{small} networks with \emph{medium} density. There are 10 networks and the model was run 3 times (with different seeds), hence the maximum number of counts per lower bound is 30. The gray area represents the bounded region given by the lower ($l$) and upper ($c$) bounds (i.e., the area in which an output should occur). The dashed line represents the ground-truth number of clusters. (b) A box-and-whisker plot of the adjusted rand index (ARI) corresponding to (a).}
	\label{fig:varying_lower_bound}
\end{figure}

Our proposed constraint is successfully enforced in our synthetic experiments.

In our first set of experiments, we asked the following questions: does the constraint work and, if so, how well? To address these questions, we applied our model (\emph{GNN+REG+CONSTRAINT}) to 10 \emph{small}-sized, \emph{medium} density networks and executed it 3 separate times, yielding 30 experiments per lower bound. From figure~\ref{fig:varying_lower_bound_1}, we see that in all but one experiment, the lower bound was successfully enforced. The scenario in which the constraint was not satisfied occurred when the lower bound ($l=7$) exceeded the ground-truth number of clusters ($k=5$). To fix the issue, we increased the constant $\mu$ (setting it to $\mu = 1000$) from equation~\ref{eqn:constraint} and thus increased the effect of the constraint. Additionally, we note from figure~\ref{fig:varying_lower_bound_1} that the output number of communities is more diverse when the lower bound is below the ground truth. However, when the lower bound equals or exceeds the ground-truth, the returned number of communities is almost exclusively on the lower bound. This is expected if the true number of communities ($k=5$) yields the highest modularity score.

The next question is: how well does the constraint work? In other words, does it produce meaningful communities? We plot the adjusted rand index (ARI) (described in equation~\ref{eqn:ari}) for the varying-lower-bound experiments in figure~\ref{fig:varying_lower_bound_2} to assess the quality of the clustering. The plot shows high ARI scores, indicating that the model found meaningful communities within the bounds of the constraint. The ARI also has a steady trend with a slight peak at 4 and 5 communities. As the lower bound increases past 5 communities (the true number of communities), the ARI decreases monotonically. This behavior is expected because, as the true number of communities is exceeded, there are fewer nodes with the correct neighbors and, as such, there will be a lower ARI score.

\begin{figure}[h!]
	\centering
	\includegraphics[width=0.9\linewidth]{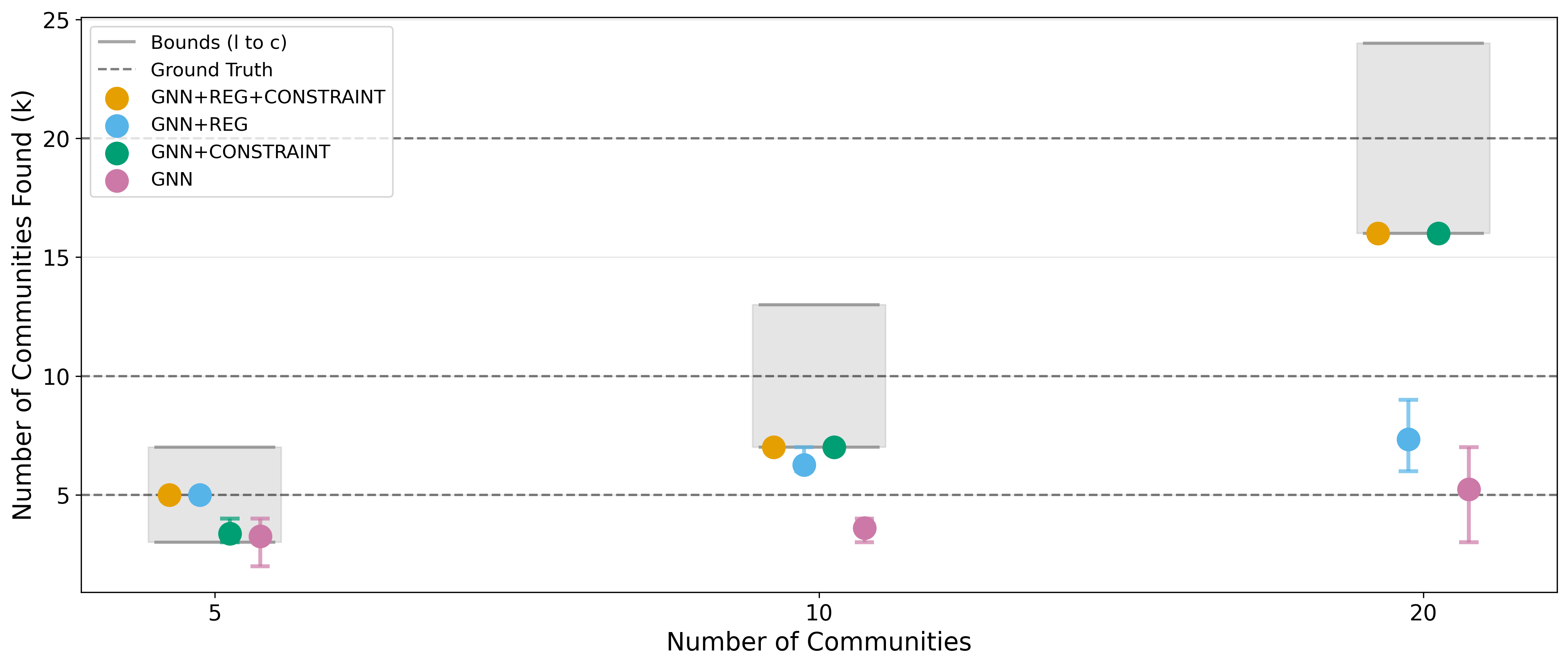}
	\caption{The number of communities found by each model when the number of clusters is varied. The experiments were performed on \emph{medium} networks with \emph{medium} density. 10 networks were generated for each value on the x-axis, and each model was run 3 separate times per network. The gray shaded area represents the bounded region for \emph{GNN+REG+CONSTRAINT} and \emph{GNN+CONSTRAINT}, and the horizontal dashed lines represent the ground-truth number of clusters. The points represent the average output for each model, with error bars indicating the minimum and maximum.}
	\label{fig:varying_cluster_number}
\end{figure}

For the next batch of experiments, we tested whether the constraint still held if the network properties were varied. We varied the network size, density, and number of true communities. In all cases, the constraint was satisfied. We examine the results of varying the number of true communities in figure~\ref{fig:varying_cluster_number}. We used 10 \emph{medium} networks with \emph{medium} density for each number of true communities. For example, there are 10 networks with \emph{medium} density that have 5 true communities. As the number of communities increases, the number of output communities remains on the lower bound for \emph{GNN+REG+CONSTRAINT} and \emph{GNN+CONSTRAINT}. On the other hand, the models without the constraint diverge further from the ground truth. This is especially evident when comparing $10$ and $20$ true communities. The models \emph{GNN} and \emph{GNN+REG} output a similar number of communities in both cases, even though there are twice the number of communities in the latter case. This is somewhat expected, given that the communities are unequal in size and randomly generated (we provide the community sizes in table~\ref{tab:cluster_sizes}). As the number of communities increases, there is a greater chance for smaller communities that are difficult to detect with modularity. Modularity is known to have a resolution limit \cite{fortunato2007resolution, good_performance_2010, fortunato_community_2016} that ultimately biases modularity maximisation approaches toward larger communities. This further illustrates the value of a lower bound in that it forced the models \emph{GNN+REG+CONSTRAINT} and \emph{GNN+CONSTRAINT} to be closer to the desired number of communities (i.e., the ground truth) and, as a result, counteracted the bias of modularity. This is an indirect consequence of the constraint that is dependent on the lower bound chosen by the user. Other methods address the resolution limit more explicitly, like \cite{reichardt2006statistical,xu2011multiple}, which introduce a hyperparameter that can be adjusted to identify smaller communities. We discuss the constraint and its relation to modularity's resolution limit further in section~\ref{subsec:resolution_limit}.

\begin{figure}[h!]
	\centering
	
	\begin{subfigure}[b]{0.48\textwidth}
		\centering
		\includegraphics[width=\textwidth]{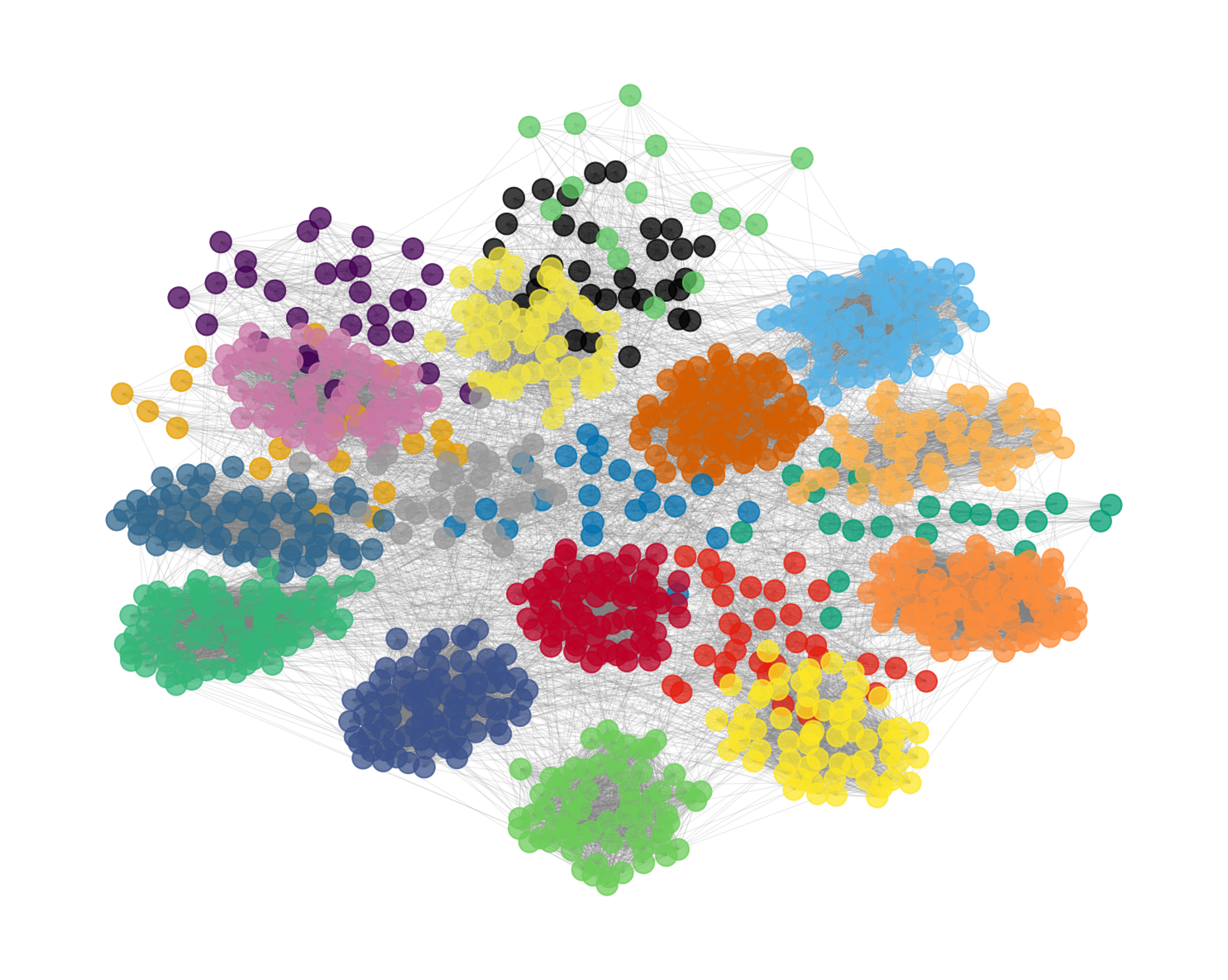}
		\caption{Ground Truth (k=20)}
		\label{fig:network_gt}
	\end{subfigure}
	\hfill
	\begin{subfigure}[b]{0.48\textwidth}
		\centering
		\includegraphics[width=\textwidth]{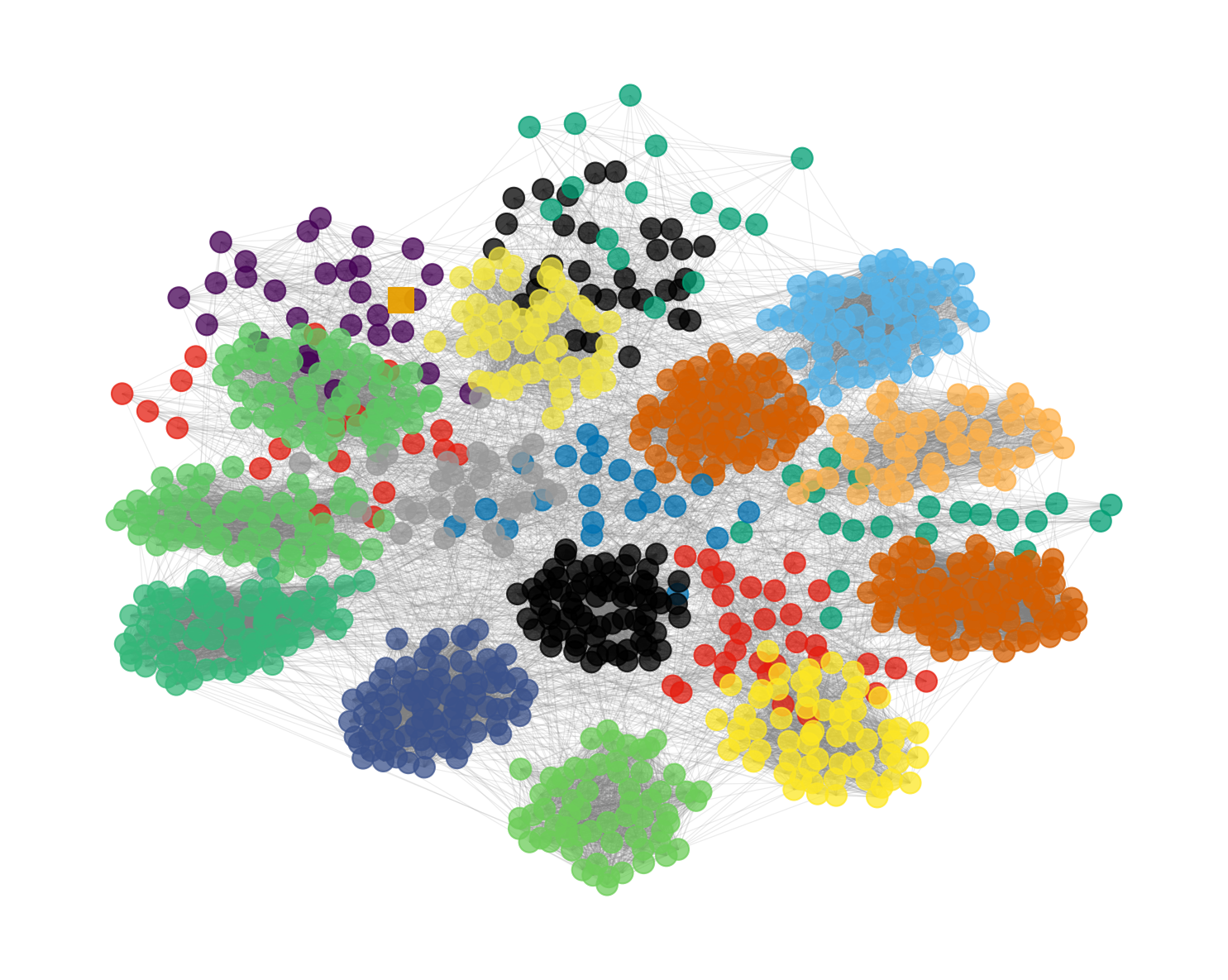}
		\caption{GNN+REG+CONSTRAINT (k=16)}
		\label{fig:network_gnn_reg_constraint}
	\end{subfigure}
	
	\vspace{0.5cm}
	
	\begin{subfigure}[b]{0.32\textwidth}
		\centering
		\includegraphics[width=\textwidth]{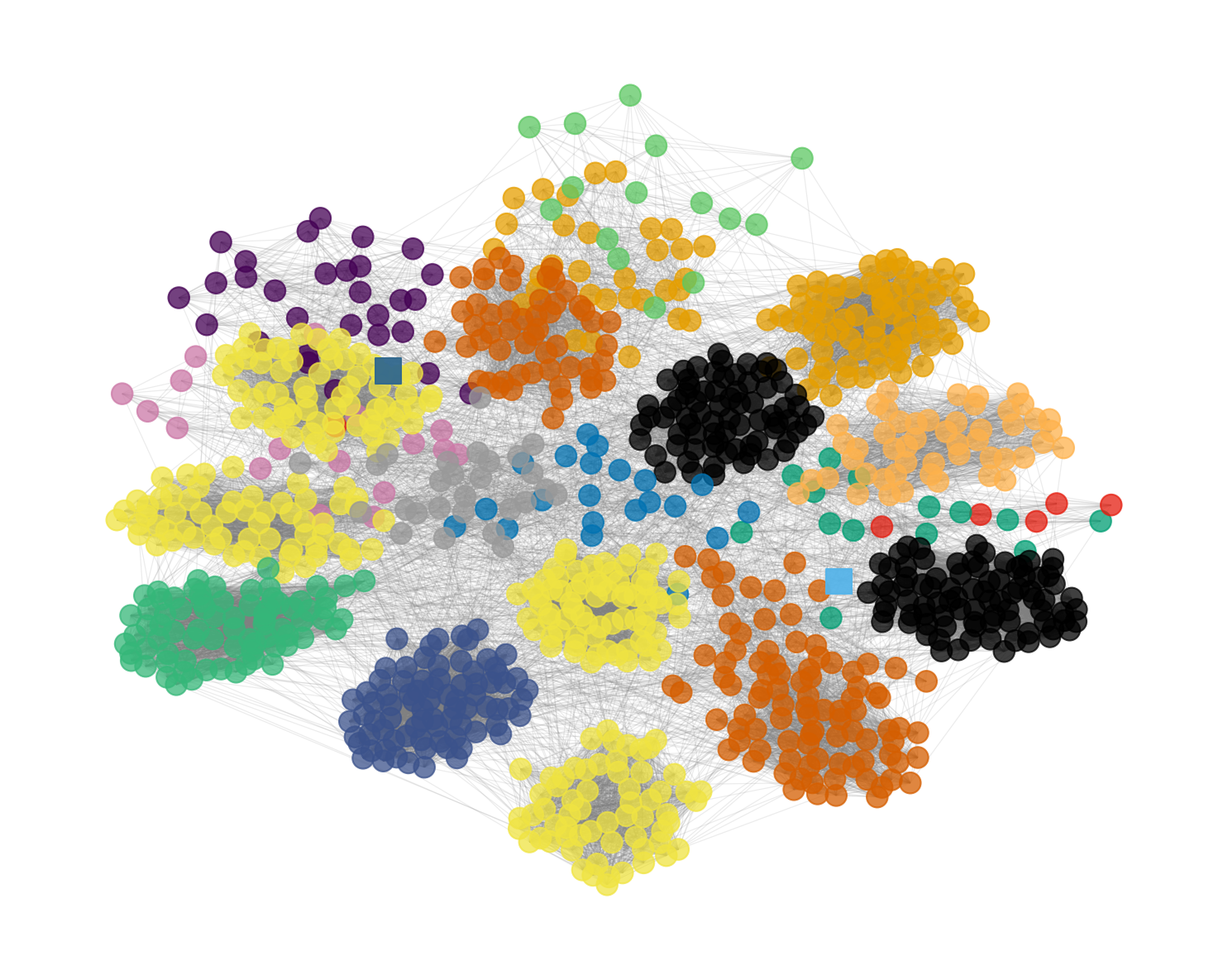}
		\caption{GNN+CONSTRAINT (k=16)}
		\label{fig:network_gnn_constraint}
	\end{subfigure}
	\hfill
	\begin{subfigure}[b]{0.32\textwidth}
		\centering
		\includegraphics[width=\textwidth]{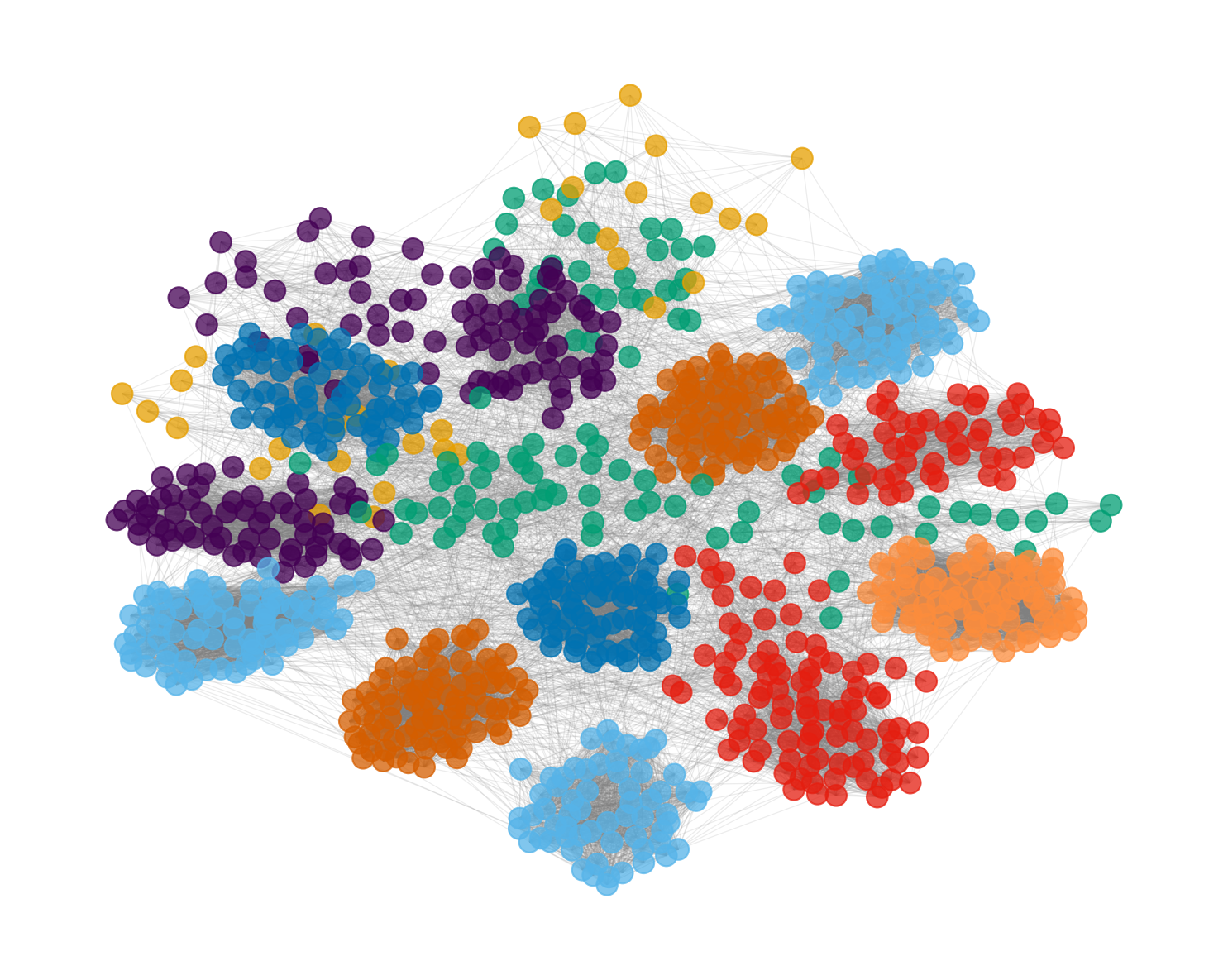}
		\caption{GNN+REG (k=8)}
		\label{fig:network_gnn_reg}
	\end{subfigure}
	\hfill
	\begin{subfigure}[b]{0.32\textwidth}
		\centering
		\includegraphics[width=\textwidth]{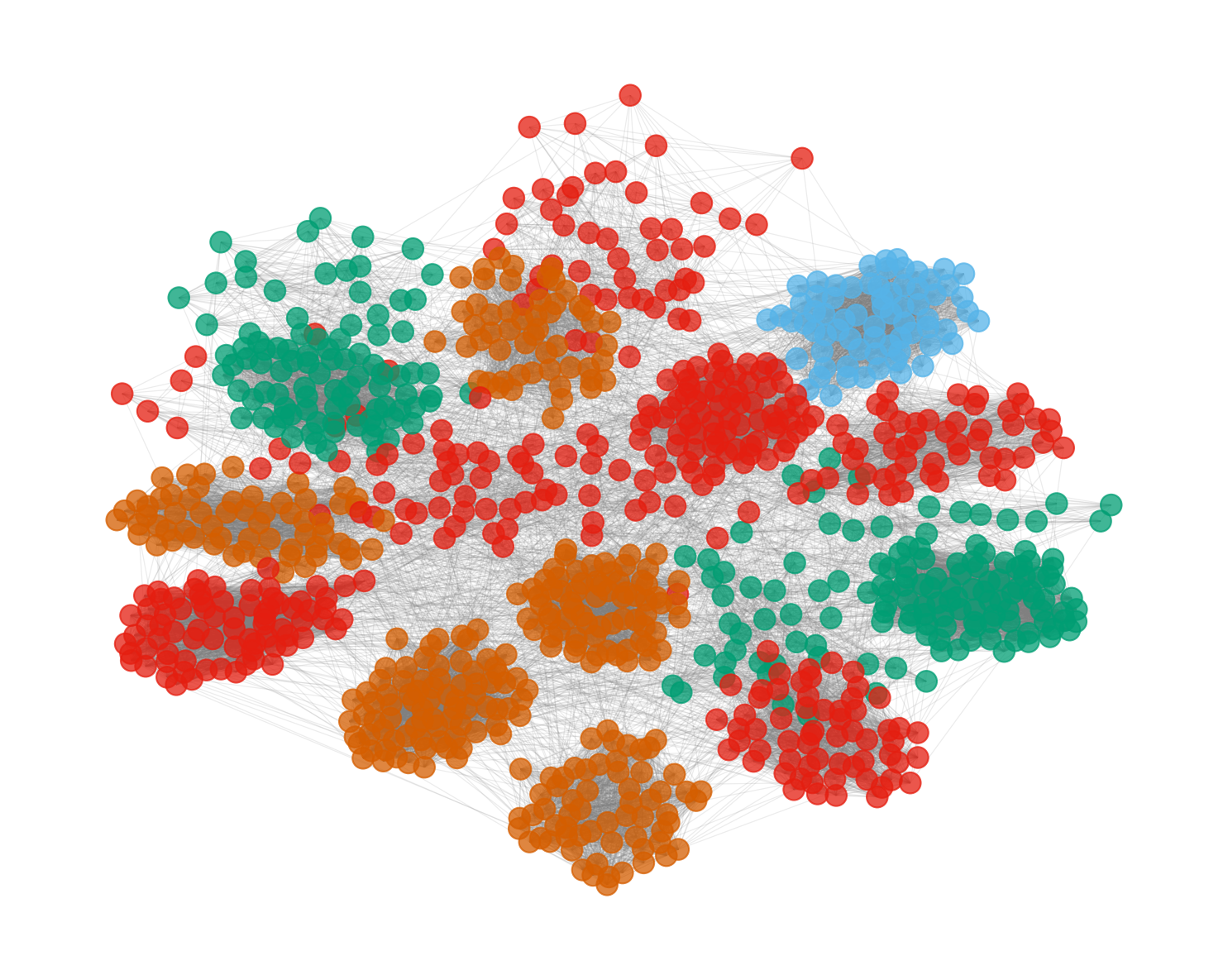}
		\caption{GNN (k=4)}
		\label{fig:network_gnn}
	\end{subfigure}
	
	\caption{Network visualizations showing community assignments for each method for one run on one \emph{medium}-sized, \emph{medium}-density network. Each sub-figure displays the same network with nodes colored according to their assigned communities. Single-node communities are shown as squares, while multi-node communities are shown as circles. There is one square in (b) and two squares in (c).}
	\label{fig:network_visualizations}
\end{figure}
\begin{figure}[h!]
	\centering
	\includegraphics[width=0.7\textwidth]{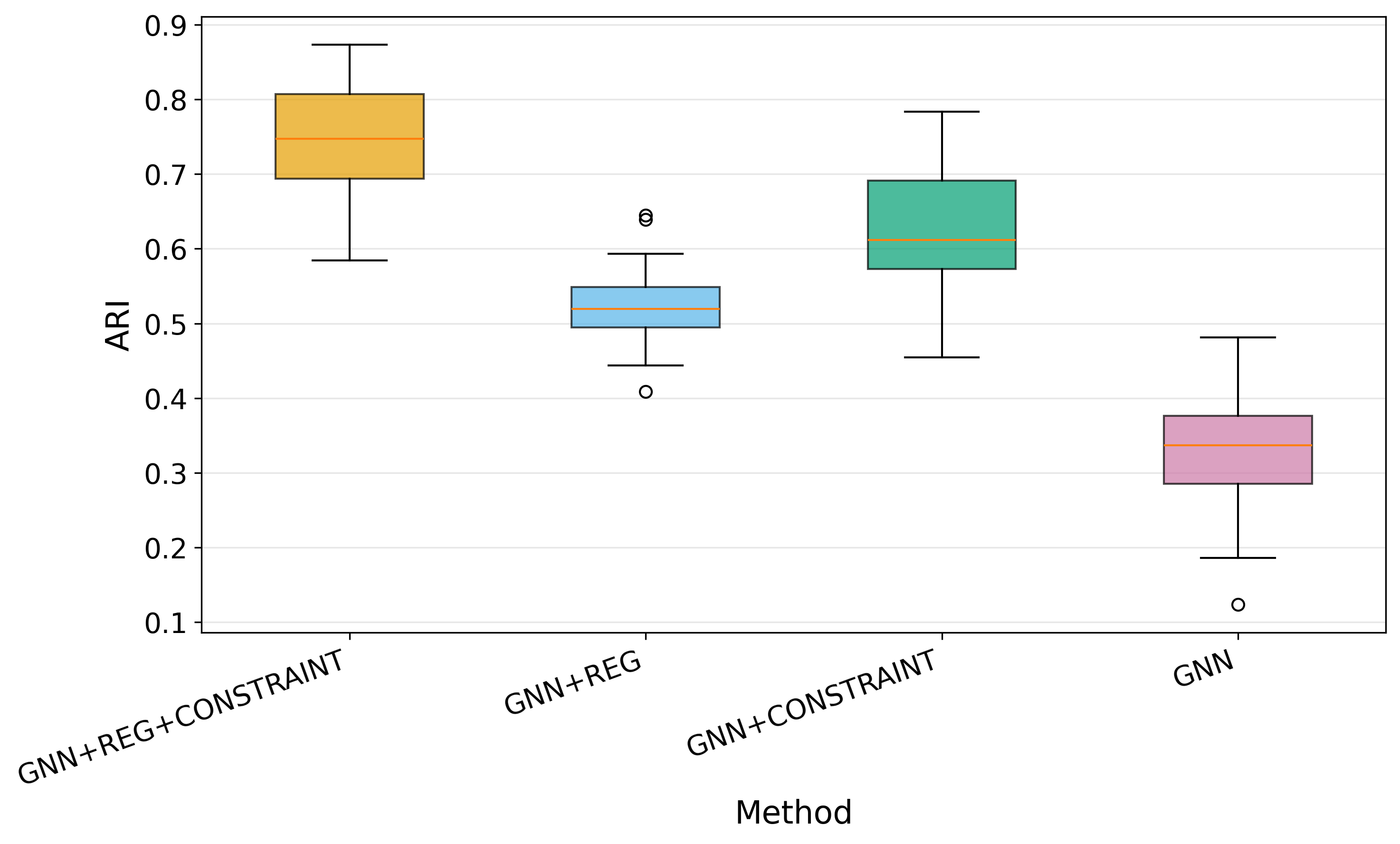}
	\caption{A box-and-whisker plot for the adjusted rand index (ARI) score of each model on 10 \emph{medium} networks with \emph{medium} density and 20 ground-truth communities. Each model was run three times, resulting in 30 experiments per model.}
	\label{fig:ari_20clusters}
\end{figure}

To elucidate the scenario of $20$ true communities in figure~\ref{fig:varying_cluster_number}, we visualize the output of each model in figure~\ref{fig:network_visualizations} and plot the ARI in figure~\ref{fig:ari_20clusters}. In figure~\ref{fig:network_visualizations}, we plot the predicted labels of each model for a single run on a single \emph{medium} network with \emph{medium} density. \emph{GNN+REG+CONSTRAINT} found 9 communities that perfectly match the ground truth; \emph{GNN+CONSTRAINT} found 7 and \emph{GNN+REG} and \emph{GNN} found 1. We see this reflected in figure~\ref{fig:ari_20clusters} as \emph{GNN+REG+CONSTRAINT} and \emph{GNN+CONSTRAINT} have higher ARI scores than \emph{GNN+REG} and \emph{GNN}. This implies that the constraint allowed the models to find more meaningful communities. Furthermore, our proposed approach, \emph{GNN+REG+CONSTRAINT}, has the highest ARI. This is also expected since the addition of the regularization term encourages more nodes per community, which would lead to a better ARI. In contrast, the constraint in \emph{GNN+CONSTRAINT} only requires one node per community.

We show the results obtained by varying the network size and density in the appendix. Furthermore, we assessed the special case where the upper bound equals the lower bound and present the results in the appendix (figure~\ref{fig:upper_lower}). Again, in all cases, the constraint was successfully enforced. 

In table~\ref{tab:runtime_syn}, we note the runtime of each model in seconds. We observe that all models have comparable runtime and, as a result, the addition of the constraint and regularization term have little effect.

\begin{table}[h!]
	\footnotesize
	\caption{Runtime (in seconds) of each model on synthetic datasets. Each entry reports the mean $\pm$ standard deviation over three runs.}
	\begin{tabular}{ccc|llll}
		\rot{Size label }& \rot{Clusters }& \rot{Density label} &\rot{GNN+REG+CONSTRAINT}	&\rot{GNN+CONSTRAINT}  				&\rot{GNN+REG}  					& \rot{GNN} \\
		\midrule
		Small 		& 5 			&Low 						&\phantom{00}7.36$\pm$\phantom{00}0.05 							& \phantom{00}7.06$\pm$\phantom{0}0.09								&\phantom{00}7.09$\pm$\phantom{00}0.16	 &\phantom{00}6.69$\pm$\phantom{00}0.08	 \\		
		&				&Medium 				&\phantom{00}7.51$\pm$\phantom{00}0.01	 							& \phantom{00}7.16$\pm$\phantom{0}0.02								&\phantom{00}7.13$\pm$\phantom{00}0.03	  &\phantom{00}6.89$\pm$\phantom{00}0.15\\		
		
		\midrule
		Medium & 5				&Low 						&\phantom{0}96.16$\pm$\phantom{00}6.03 						&\phantom{0}70.93$\pm$14.80 							&\phantom{0}76.07$\pm$\phantom{00}2.93 &\phantom{0}76.21$\pm$\phantom{0}11.57\\
		& 				&Medium					&316.26$\pm$\phantom{0}10.58 						& 318.85$\pm$\phantom{0}8.53							&210.70$\pm$\phantom{0}94.60	 &238.49$\pm$\phantom{0}57.46\\
		& 				&High						&329.22$\pm$141.87	 					& 526.20$\pm$\phantom{0}0.69							&377.39$\pm$118.24 &342.75$\pm$148.17\\
		
		&10 		&Medium					&209.06$\pm$\phantom{0}45.72						& 149.10$\pm$62.83								&214.92$\pm$\phantom{0}27.75&192.93$\pm$\phantom{0}39.13\\
		
		&20 		&Medium					&125.86$\pm$\phantom{0}35.65						&  143.05$\pm$25.50								&109.57$\pm$\phantom{0}34.71&122.12$\pm$\phantom{0}21.76\\
		
		\midrule
		Large 		& 5			&Low 						&33 935.11$\pm$6 138.76				&34 362.55$\pm$9 141.21						&32 890.65$\pm$11 032.64  &22 623.14$\pm$7 883.28\\

		\bottomrule
	\end{tabular}
	\label{tab:runtime_syn}
\end{table}

Lastly, the four models, each using a different loss function, help assess the influence of the balance and constraint terms separately. We observe that the constraint is enforced when it is the only addition to the \emph{GNN}, confirming that the constraint's performance is independent of the balance-cluster regularization term. However, the constraint requires only that each community contains at least one node, which is not optimal for modularity. Based on our experiments, the modularity term is not enough to encourage more nodes per community, as may be seen in figure~\ref{fig:varying_cluster_number} when comparing the \emph{GNN+CONSTRAINT} and \emph{GNN+REG+CONSTRAINT} for \textit{5} ground-truth communities. The \emph{GNN+CONSTRAINT} results remain on the lower bound while the  \emph{GNN+REG+CONSTRAINT} results consistently find the ground truth. 
Moreover, the addition of only the balance term to \emph{GNN} also seems to help the model find more communities, illustrated again in figure~\ref{fig:varying_cluster_number} when comparing the models \emph{GNN} and \emph{GNN+REG}. \emph{GNN+REG} is consistently closer than \emph{GNN} to the true number of communities. Hence, this implies that the balance term encourages more nodes per community and is effective even in cases of unbalanced communities, ultimately leading to a higher-quality clustering.

It is important to note that ARI can only be assessed when the true communities are known (typically in synthetic data). In contrast, we rarely know the true communities in real data and, as such, the use of metrics (such as ARI) is no longer helpful or applicable.

\subsection{Results on real data}

We use a collection of widely studied real-world graph datasets, summarized in table~\ref{tab:real_stats}.
Cora, Citeseer, and PubMed are citation networks in which nodes correspond to scientific publications, edges represent citation links, node features are bag-of-words representations of document content, and communities are given by the topic labels of the papers (e.g., seven classes in Cora, six in Citeseer, and three in PubMed)  \cite{sen2008collective}.
The Actor dataset is a co-occurrence network in which nodes are actors and edges indicate co-occurrence in films; node features are derived from textual attributes, and communities correspond to actor or role categories \cite{pei2020geomgcn}.
In all cases, the real data come with labels. These are often used for supervised learning tasks but may not necessarily be linked to the community structure found in the network, as in real networks where the ground truth is unknown. For our purposes, the labels serve as an arbitrary reference point for the bounds, instead of a true or accurate number of communities that we could use to assess our constraint.

\begin{table}[h!]
	\renewcommand*{\arraystretch}{1.2}
	\caption{A summary of the real-network statistics.}
	\centering
	\begin{tabular}{ccc}
		\toprule
		Dataset  & Nodes 	& Edges 	\\
		\midrule
		PubMed 	&19 717			& 44 338  \\
		\midrule
		Actor 	& 7 600			&30 019  \\
		\midrule
		Citeseer 	& 3 312			& 4 715	\\
		\midrule
		Cora 		& 2 708			&5 429   \\
		\bottomrule
	\end{tabular}
	\label{tab:real_stats}
\end{table}

\begin{figure}[h!]
	\centering
	
	\begin{subfigure}[b]{0.455\textwidth}
		\centering
		\includegraphics[width=1.1\textwidth]{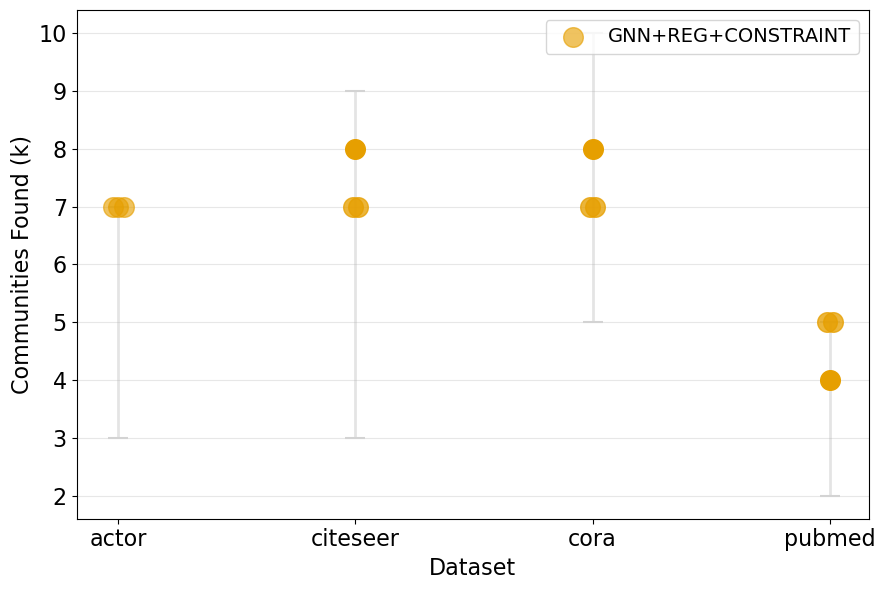}
		\caption{}
		\label{fig:real_data_bounds}
	\end{subfigure}
	\hfill
	\begin{subfigure}[b]{0.455\textwidth}
		\centering
		\includegraphics[width=1.1\textwidth]{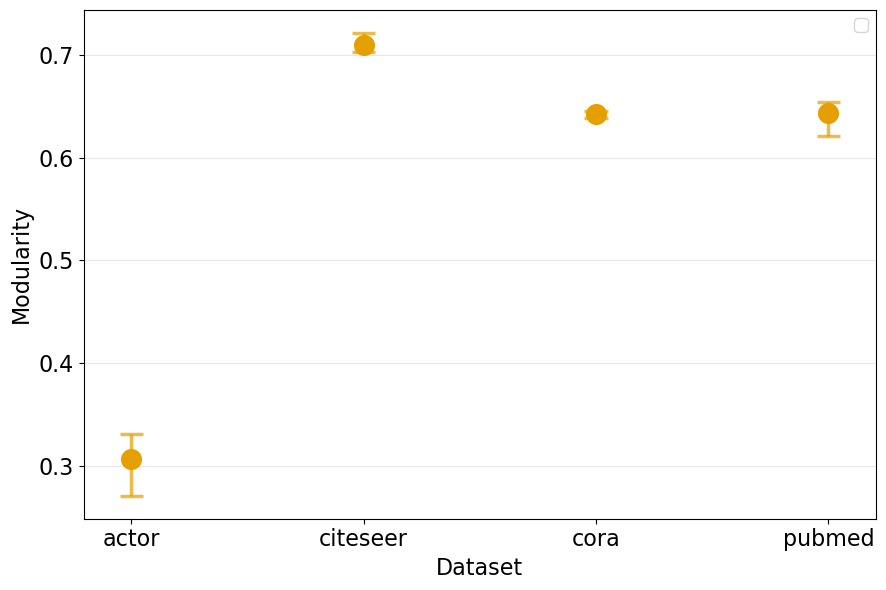}
		\caption{}
		\label{fig:real_data_modularity}
	\end{subfigure}
	
	\caption{(a) The number of communities predicted for the real datasets (table~\ref{tab:real_stats}). The vertical gray bars represent the constrained region designated by the lower ($l$) and upper ($c$) bounds. The model was run three times with different seeds, hence there are three points per dataset. (b) The modularity values corresponding to (a). The points represent the mean, and the error bars signify the minimum and maximum.}
	\label{fig:real_data}
\end{figure}

Similar to the synthetic networks, the constraint is enforced in all cases. Figure~\ref{fig:real_data_bounds} depicts the predicted number of communities for each dataset using the model \emph{GNN+REG+CONSTRAINT}, where the gray vertical bars represent the bounded region demarcated by the upper and lower bounds. All points lie within the bounds. Figure~\ref{fig:real_data_modularity} depicts the associated mean modularity output by the model for each dataset. The model yields high modularity for the Citeseer, Cora, and PubMed datasets, and low modularity for the Actor dataset. Notably, even though the model output a low modularity, it is not a result of the constraint. Table~\ref{tab:modularity_real} depicts the modularity for all models (with and without the constraint), each of which converged to a low modularity. This may be a result of poor performance of the GNNs, or the dataset has a poor community structure. This highlights the difficulty of evaluating models on real data where the ground truth is unknown.

\begin{figure}[h!]
	\centering
	
	\begin{subfigure}[b]{0.48\textwidth}
		\centering
		\includegraphics[width=\textwidth]{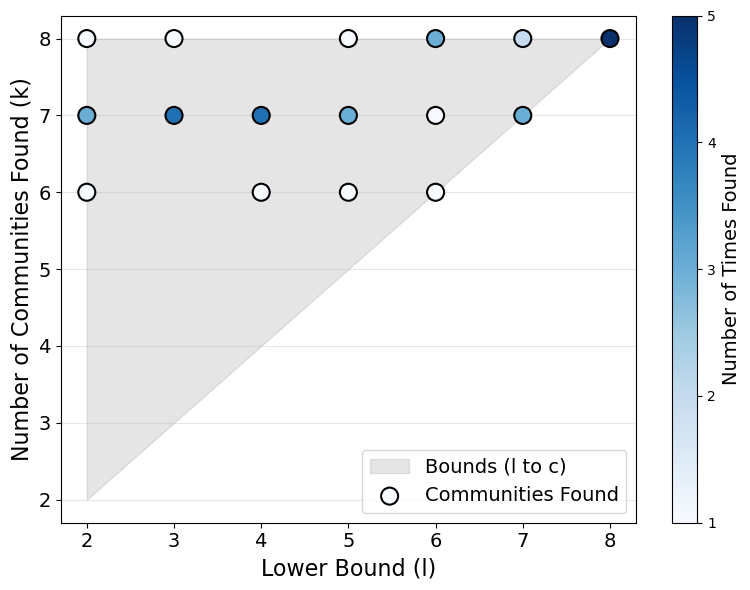}
		\caption{}
		\label{fig:cora_varying_lower_bound_1}
	\end{subfigure}
	\hfill
	\begin{subfigure}[b]{0.48\textwidth}
		\centering
		\includegraphics[width=\textwidth]{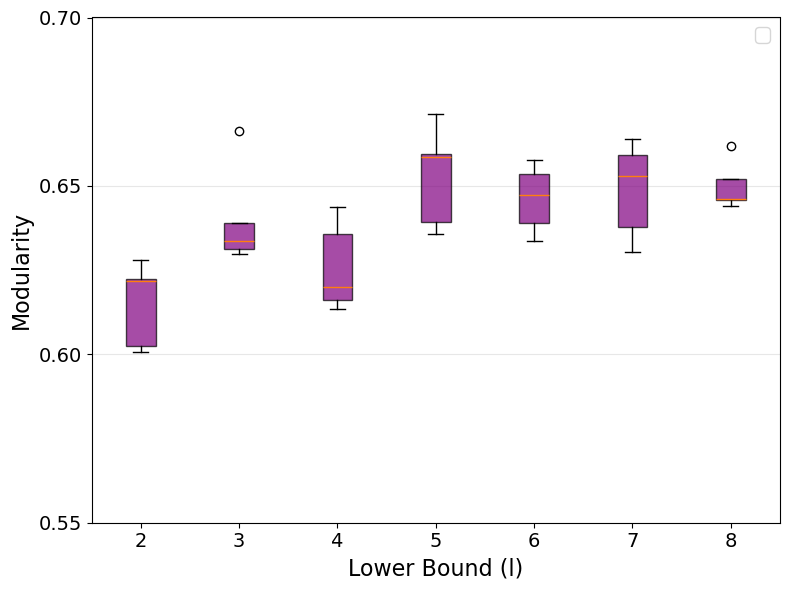}
		\caption{}
		\label{fig:cora_varying_lower_bound_2}
	\end{subfigure}
	
	\caption{(a) The number of communities predicted by model \emph{GNN+REG+CONSTRAINT} as the lower bound varies on the Cora dataset. The model was run 5 times (with different seeds), hence there are a maximum of 5 counts per lower bound. The gray area represents the bounded region given by the lower ($l$) and upper ($c$) bounds (i.e., the region in which an output should occur). (b) A box-and-whisker plot of the modularity corresponding to (a).}
	\label{fig:cora_varying_lower_bound}
\end{figure}

In addition, figure~\ref{fig:cora_varying_lower_bound} illustrates the result of varying the lower bound on the Cora dataset. Again, the constraint held in each case (figure~\ref{fig:cora_varying_lower_bound_1}). Interestingly, the modularity improved (figure~\ref{fig:cora_varying_lower_bound_2}) as the bounds tightened.
In table~\ref{tab:runtime_real}, we show the runtime for each model in seconds. We note that, as with the synthetic data, the addition of the constraint and balance regularization term has no meaningful effect on the runtime.

\begin{table}[h!]
	\centering
	\renewcommand*{\arraystretch}{1.2}
	\caption{The runtime (in seconds) for real data. Each entry reports the mean $\pm$ standard deviation over three runs.}
	\begin{tabular}{c|cccc}
		\toprule
		&Cora&Citeseer&PubMed&Actor\\
		\hline
		GNN+REG+CONSTRAINT	&198 $\pm$24	&464 $\pm$38	& 5 248 $\pm$\phantom{0}922 & 1 116 $\pm$180\\
		GNN+CONSTRAINT			&215 $\pm$19	&530 $\pm$45	& 6 403 $\pm$\phantom{0}294& 1 325 $\pm$\phantom{0}71\\
		GNN+REG							&223 $\pm$23	&555 $\pm$37	& 6 123 $\pm$\phantom{0}602& 1 290 $\pm$109\\
		GNN										&218 $\pm$21	&492 $\pm$83	& 5 304 $\pm$1659& 1 201 $\pm$216\\
		\bottomrule
	\end{tabular}
	\label{tab:runtime_real}
\end{table}

\begin{figure}[htbp]
	\centering
	\begin{subfigure}[b]{0.23\textwidth}
		\centering
		\includegraphics[width=\textwidth]{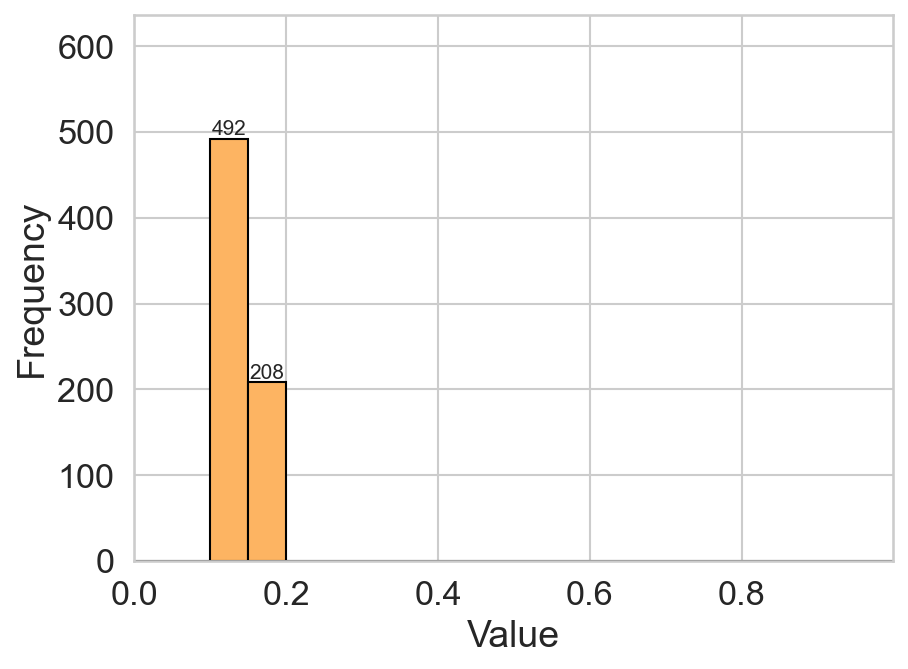}
		\caption{Epoch 1}
		\label{fig:hist_epoch1}
	\end{subfigure}
	\hfill
	\begin{subfigure}[b]{0.23\textwidth}
		\centering
		\includegraphics[width=\textwidth]{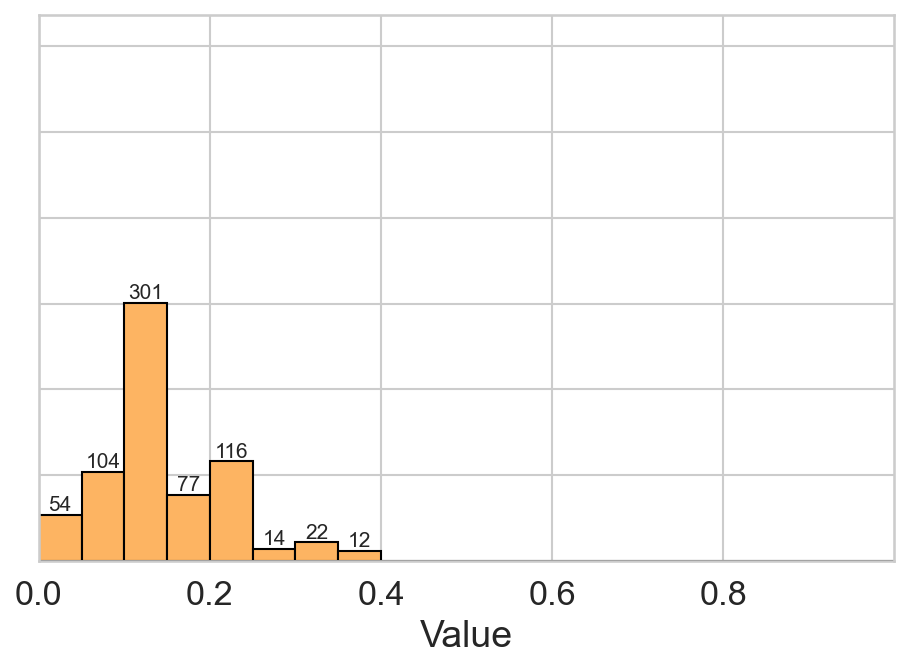}
		\caption{Epoch 51}
		\label{fig:hist_epoch51}
	\end{subfigure}
	\hfill
	\begin{subfigure}[b]{0.23\textwidth}
		\centering
		\includegraphics[width=\textwidth]{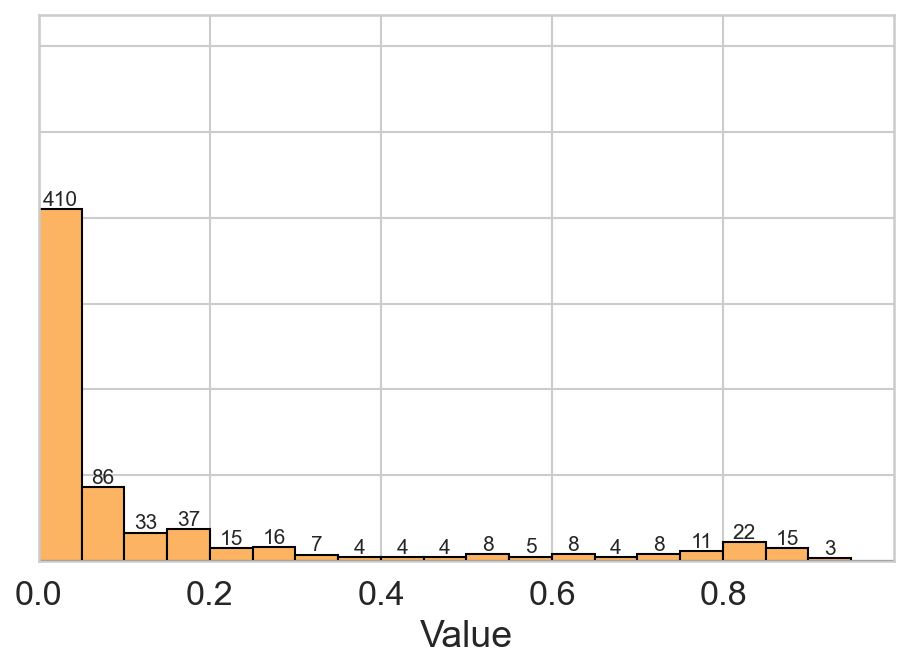}
		\caption{Epoch 101}
		\label{fig:hist_epoch101}
	\end{subfigure}
	\hfill
	\begin{subfigure}[b]{0.23\textwidth}
		\centering
		\includegraphics[width=\textwidth]{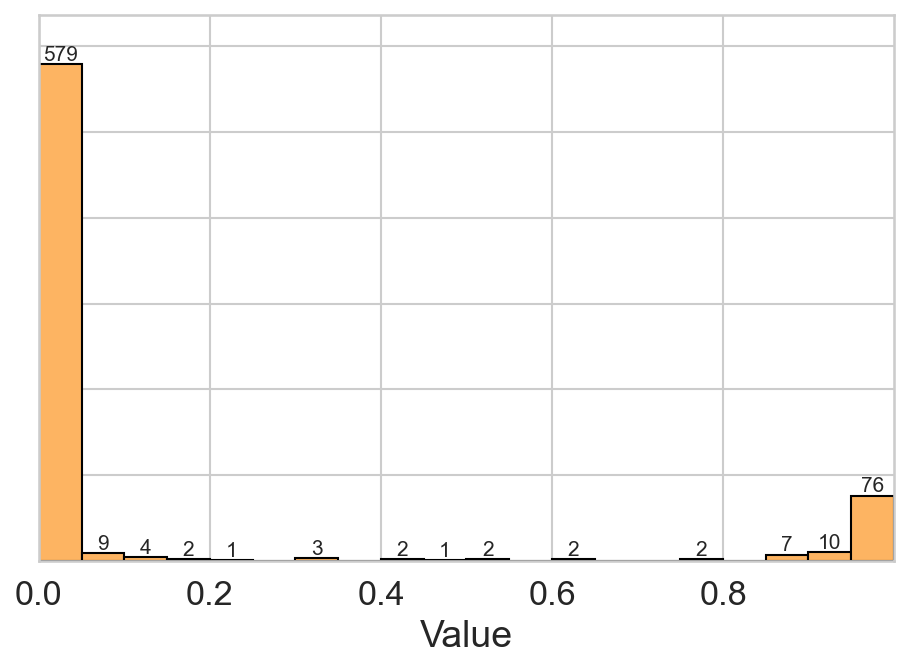}
		\caption{Epoch 200}
		\label{fig:hist_epoch200}
	\end{subfigure}
	\caption{Example distribution of cluster assignment values during training for 200 epochs. The cluster assignment evolves from (a) -- a random initialization of the cluster assignment -- to (d). Training was performed with \emph{GNN+REG+CONSTRAINT} on a \emph{small} network with \emph{medium} density.}
	\label{fig:histograms_all}
\end{figure}

Finally, from a more philosophical perspective, the constraint interprets the cluster assignment matrix as an accurate representation of the model's certainty of the cluster assignment. In other words, the constraint does not consider the quality of the assignment with respect to the community quality function. In practice, the cluster assignment matrix is not an accurate representation of the model's certainty about the cluster assignments (consider that the cluster assignment matrix is initialized with random weights). So, initially, the model guesses the cluster assignment. Instead, the cluster assignment matrix can be viewed as the set of values that maximizes the loss function. Figure \ref{fig:histograms_all} illustrates an example of the evolution of the cluster assignment during training. In practice, this difference in perspective appears to have no effect on the constraint's efficacy.

\section{Conclusion, limitations, and future work}
We proposed a constraint that allows a user to specify a desired number of output communities, or a range, when using graph neural networks for community detection. Such functionality did not exist previously, as graph neural networks would not return the specified number of communities. We empirically tested our constraint on a variety of real and synthetic networks, and it was successfully enforced. We recommend using the constraint jointly with a regularization term that encourages balanced communities, or another term that would encourage multiple nodes per community. This is because the constraint is satisfied if there is one node per community. 

A limitation of our work was that all synthetic networks had a strong community structure. A promising direction would be to systematically identify the phase transition at which performance begins to deteriorate as community structure weakens, and to investigate whether this transition can be related to structural properties of the underlying network. Finally, the introduction of a constraint naturally opens the possibility to search for an optimal number of communities. The addition of such functionality to graph neural networks would be a beneficial and welcome avenue for future work.

%
%


\funding{This work was supported by the Deutsche Forschungsgemeinschaft (DFG, German Research Foundation) under Germany’s Excellence Strategy – EXC number 2064/1 – Project number 390727645. D.B. acknowledges support from ERC Starting Grant no. 101165497.}


\data{The real-network data used in our work is available at \cite{pyg_data}; the synthetic data is available at \cite{syn_data}}


\bibliographystyle{plain}
\bibliography{references}

\newpage
\appendix 

\renewcommand{\thesection}{}

\renewcommand{\thesubsection}{A.\arabic{subsection}}

\renewcommand{\thefigure}{A.\arabic{figure}}
\renewcommand{\thetable}{A.\arabic{table}}
\renewcommand{\theequation}{A.\arabic{equation}}

\setcounter{figure}{0}
\setcounter{table}{0}
\setcounter{equation}{0}

\section*{\centering Appendix}

\subsection{Additional results on synthetic data}

This section presents additional experimental results on synthetic networks that further validate the effectiveness of our constraint. We examine how the model performs under varying network densities and sizes, and explore the behavior when the upper and lower bounds are equal, effectively constraining the model to a fixed number of communities.

\begin{figure}[h!]
	\centering
	\includegraphics[width=0.8\linewidth]{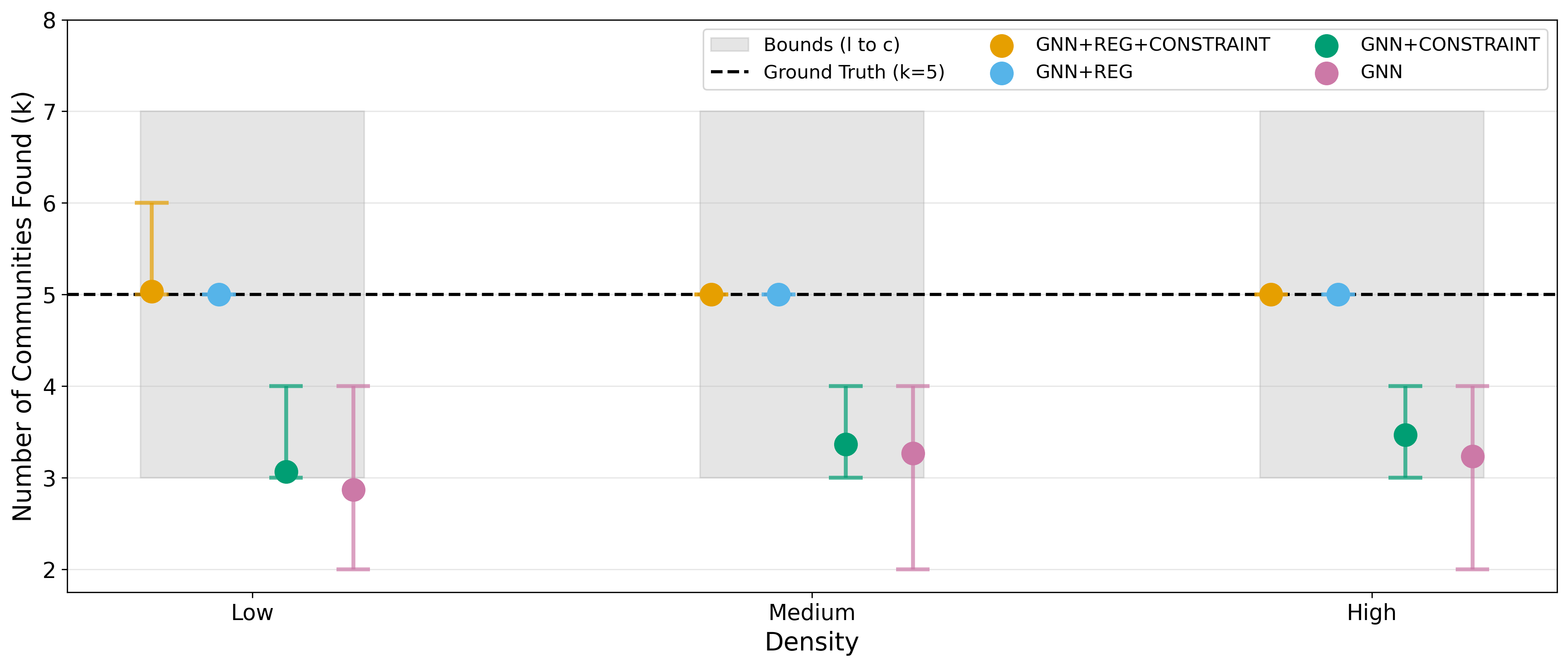}
	\caption{Number of communities predicted for 10 \emph{medium} networks with 3 different densities: \emph{low}, \emph{medium} and \emph{high}. The gray shaded region represents the range given by the lower ($l$) and upper ($c$) bounds and enforced by our constraint. The dashed line is the true number of communities ($k=5$), and the points represent the average output for each model with minimum and maximum error bars.}
	\label{fig:varying_density}
\end{figure}

Figure~\ref{fig:varying_density} demonstrates the robustness of our method across different network densities. The model maintains consistent performance in predicting the number of communities while respecting the specified bounds, regardless of whether the network has \emph{low}, \emph{medium}, or \emph{high} edge density. The balance-cluster regularization term encouraged more clusters, resulting in \emph{GNN+REG} finding the true number of communities. Thus, the constraint guaranteed that the returned number of clusters was within the bounds, and the balance regularization encouraged more clusters. This supports the use of the constraint and regularization terms in tandem.

\begin{figure}[h!]
	\centering
	\includegraphics[width=0.8\linewidth]{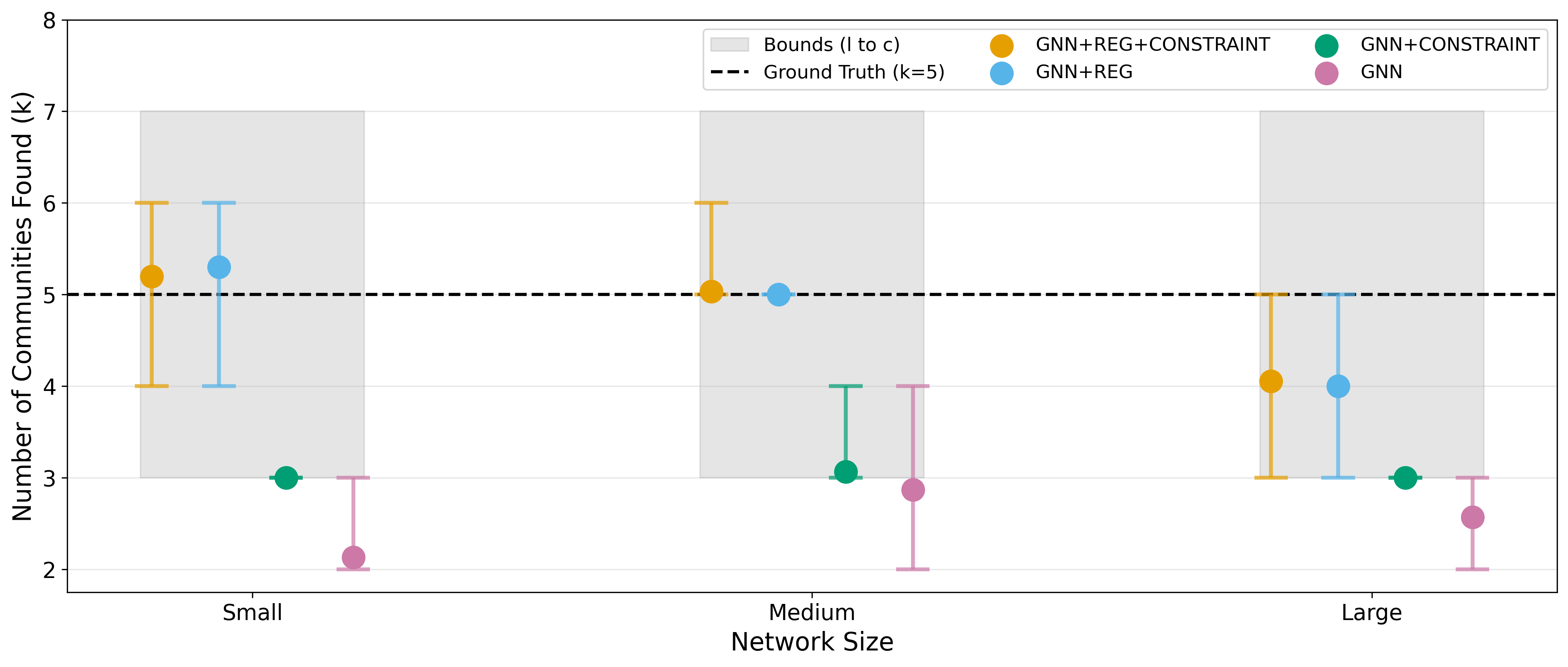}
	\caption{Number of communities found in 10 \emph{low} density networks with varying sizes of \emph{small}, \emph{medium} and \emph{large}. The gray shaded area represents the region demarcated by the lower ($l$) and upper ($c$) bounds. The dashed horizontal line represents the true number of communities ($k=5$), and the points represent the average model output with maximum and minimum error bars. Each model was run 3 times per network.}
	\label{fig:varying_network_size}
\end{figure}

We varied the network size and found similar results illustrated in figure~\ref{fig:varying_network_size}. However, the results for the \emph{small} and \emph{large} networks are more varied for \emph{GNN+REG} and \emph{GNN+REG+CONSTRAINT} than the \emph{medium} network. Moreover, the results from the \emph{large} network are further from the ground truth. Both the \emph{small} and \emph{large} networks have weaker community structure (illustrated by lower modularity values presented in table~\ref{tab:syn_stats_additional}) that could account for the larger spread of results. A weaker community structure would imply that it is more difficult to delineate communities, resulting in a wider range of results. In the case of the \emph{large} networks, the difference in size between communities is greater (table~\ref{tab:cluster_sizes}) and, as such, it is more challenging to identify the smaller communities (refer to \cite{good_performance_2010} for an explanation of the resolution limits of modularity maximization). Hence, the output is (on average) lower than the ground truth.

\begin{figure}[h!]
	\centering
	\includegraphics[width=0.7\linewidth]{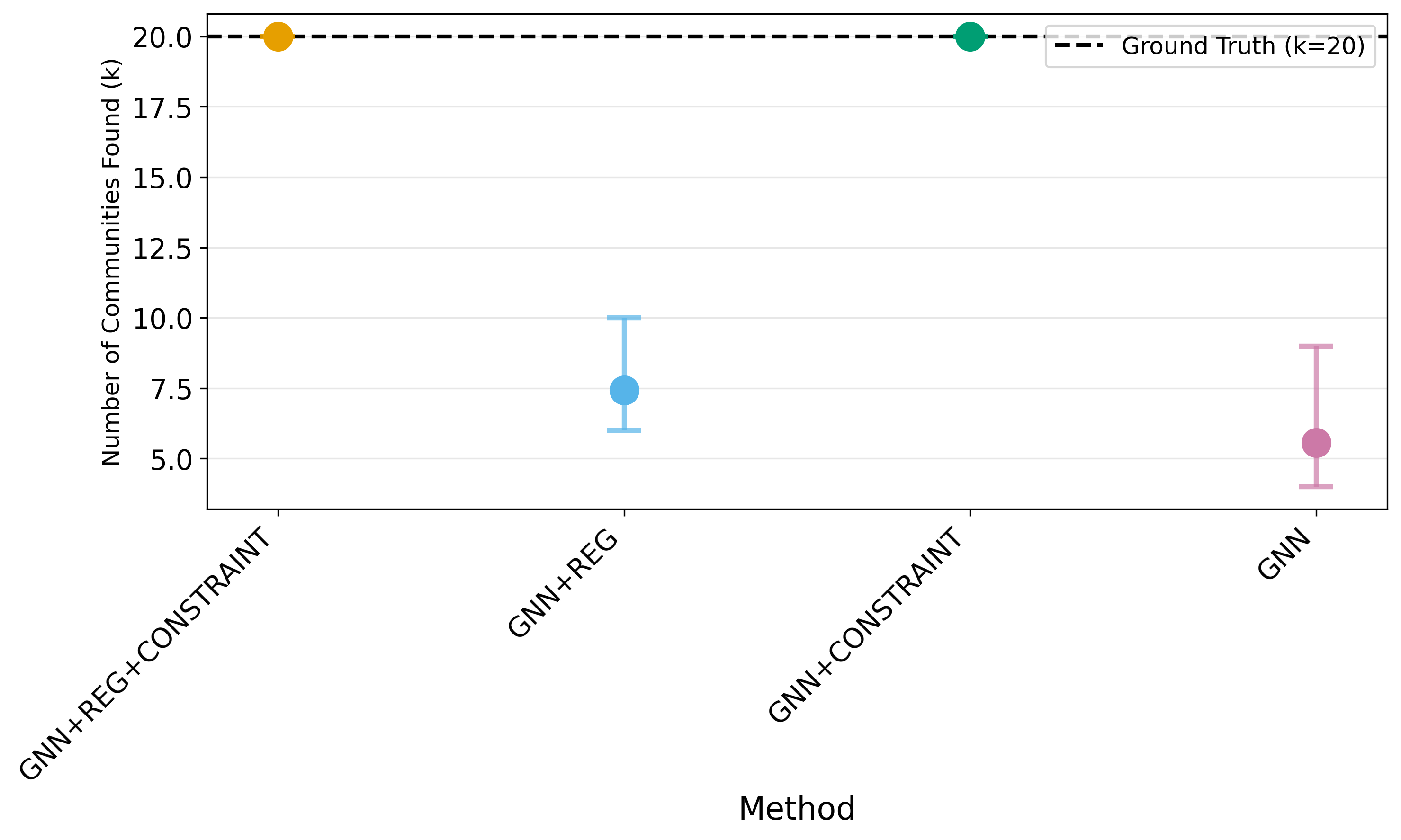}
	\caption{The number of communities predicted by the model when the lower ($l$) and upper ($c$) bounds are equal. 10 \emph{medium} networks with $20$ true communities were used, and each model was run 3 times with different seeds, resulting in 30 experiments per model. The points are the average output of the model with minimum and maximum error bars.}
	\label{fig:upper_lower}
\end{figure}

When the upper and lower bounds are equal, the constraint forces the model to predict an exact number of communities. Figure~\ref{fig:upper_lower} shows that the models with the constraint (\emph{GNN+CONSTRAINT} and \emph{GNN+REG+CONSTRAINT}) successfully adhere to the tight bound. Without the constraint, the models' output is significantly further from the true number of communities.

\subsection{Additional results on real data}

We provide extra results from our experiments on real data.

\begin{table}[h!]
	\centering
	\renewcommand*{\arraystretch}{1.2}
	\caption{The modularity for each real dataset. Each entry reports the mean $\pm$ standard deviation over three runs. Note that a standard deviation of zero indicates a value smaller than 3 decimal places.}
	\begin{tabular}{c|cccc}
		
		&Cora&Citeseer&PubMed&Actor\\
		\hline
		GNN+REG+CONSTRAINT	&0.642 $\pm$0.003	&0.710 $\pm$0.008	&0.643 $\pm$0.015 &0.306 $\pm$0.026\\
		GNN+CONSTRAINT	& 0.517 $\pm$0.017	& 0.580 $\pm$0.022	& 0.429 $\pm$0.202 & 0.294 $\pm$0.076\\
		GNN+REG	& 0.629 $\pm$0.007	& 0.703 $\pm$0.030	&0.624 $\pm$0.000 & 0.274 $\pm$0.016\\
		GNN	& 0.474 $\pm$0.019	& 0.552 $\pm$0.010	&0.200 $\pm$0.000 & 0.296 $\pm$0.069\\
		
	\end{tabular}
	\label{tab:modularity_real}
\end{table}
\subsection{Differentiability of constraint}

Our approach relies on minimizing the constraint (equation~\ref{eqn:constraint}) and, as such, differentiability is a key consideration. The derivatives of the constraint with respect to the values of the cluster assignment matrix are of concern. In other words, how does the value of the constraint change as the cluster assignment changes? In this regard, the constraint depends upon three functions that are worth further explanation: \textit{max}, \textit{sort}, and \textit{index slicing} (i.e., retrieving the top $l$ elements). Notably, we use the Python package \emph{PyTorch} version 2.2.2 \cite{paszke2019pytorch} for our calculations and, subsequently, the behavior of each derivative should be understood in that context. \textit{max} is a constant function that outputs the largest number in a set. Its derivative is zero except at the largest number, where it is one. Next, \textit{sort} arranges the elements in a set, in our case, from largest to smallest, and it has a derivative of 1 everywhere. Lastly, \textit{index slicing} retrieves $l$ elements and keeps the derivatives associated with the indexed values. 

The derivatives are calculated using the chain rule, and thus, \textit{max}, \textit{sort}, and \textit{index slicing} do not influence the magnitude of the derivatives (consider that \textit{max} and \textit{sort} have derivatives of 1, and \textit{index slicing} only restricts the derivatives available). Instead, specifically \textit{max} and \textit{index slicing}, limit the cluster assignment values that have an influence on the constraint. The part of the constraint that influences the magnitude of the derivative is $\mathbf{s}_{ik}/ \max_{1 \le k \le c} \mathbf{s}_{ik}$. Therefore, the constraint is differentiable with respect to the elements of the cluster assignment matrix, but only a few entries influence it.

Another important point to consider is when there are multiple, identical maximum entries in the \textit{max} function. In such an event, the first maximum value is output. Notably, it is a rare occurrence because the cluster assignment matrix contains real numbers between 0 and 1. For two values to be identical, they would need to be the same up to 4 decimal places.

\subsection{Alternative constraint}
Our proposed constraint (equation~\ref{eqn:constraint}) encourages that there is at least one node per cluster. A natural extension of this would be the option to include $b$ minimum number of nodes per cluster. We express this upgraded functionality in the algorithm~\ref{alg:min_k_min_size_constraint}.

\begin{algorithm}[ht]
	\caption{Constraint for finding the minimum number of clusters $l$ with a minimum number of elements $b$ per cluster -- $f(\bold{S},l,b)$}
	\label{alg:min_k_min_size_constraint}
	
	\KwIn{The cluster assignment matrix {$\bold{S} $}, the desired lower bound $l$ and the minimum number of elements per cluster $b$}
	
	\KwOut{Difference between $l\cdot b$ and $p\in \mathbb{R}_+$, the predicted number of clusters. Note that $p\leq l\cdot b$}
	
	Normalize rows of $\bold{S}$ by the largest element of each row\;
	
	Find the $b$ largest elements in each column of the row-normalized $\bold{S}$\;
	
	Find the $l$ largest row elements among the $b$ largest column elements\;
	
	$p \gets$ sum collection of elements\;
	
	\Return $l\cdot b - p$\;
	
\end{algorithm}

\subsection{User guide for constraint}

The constraint is most effective when used in tandem with a regularization term that encourages multiple nodes per cluster, such as the balance-cluster regularization term (equation \ref{eqn:balance_reg}). Without such regularization, the constraint will be satisfied by assigning a single node to a community. 

The constraint has a range of 1 to $l$ (the lower bound) which is generally much larger than the balance-regularization term, which is between 0 and 1. This naturally leads the constraint to have more of an influence (since we would like to enforce a lower bound). However, the hyperparameters  $\lambda$ and $\mu$ allow for the strength of the regularization term and constraint to be adjusted. By default, we set $\lambda=\mu=1$, but they may be altered by the user depending on the desired outcome. For example, if an output community has a single node, the value of $\lambda$ should be increased. In the case where the constraint is not satisfied, $\mu$ should be increased.

Notably, the effectiveness of the constraint fundamentally depends on the quality of the cluster assignment matrix. When the cluster assignment values are very similar (often an indication of weak partitioning between clusters), satisfying the constraint becomes challenging. In such scenarios, the model may struggle to find assignments that simultaneously respect the constraint and maintain reasonable clustering quality. In such a case, we would recommend examining the similarity between values in the same row of the cluster assignment matrix, especially the two largest values. Then, increase the constant $\mu$ (equation~\ref{eqn:loss}) that influences the strength of the constraint such that it is greater than the smallest difference between the two largest values in every row. The increase of $\mu$ might need to be orders of magnitude greater than the aforementioned difference.

\subsection{Hyperparameter sensitivity}\label{subsec:hyperparameter_sensitivity}
\begin{figure}[h!]
	\centering
	
	\begin{subfigure}[b]{0.48\textwidth}
		\centering
		\includegraphics[width=\textwidth]{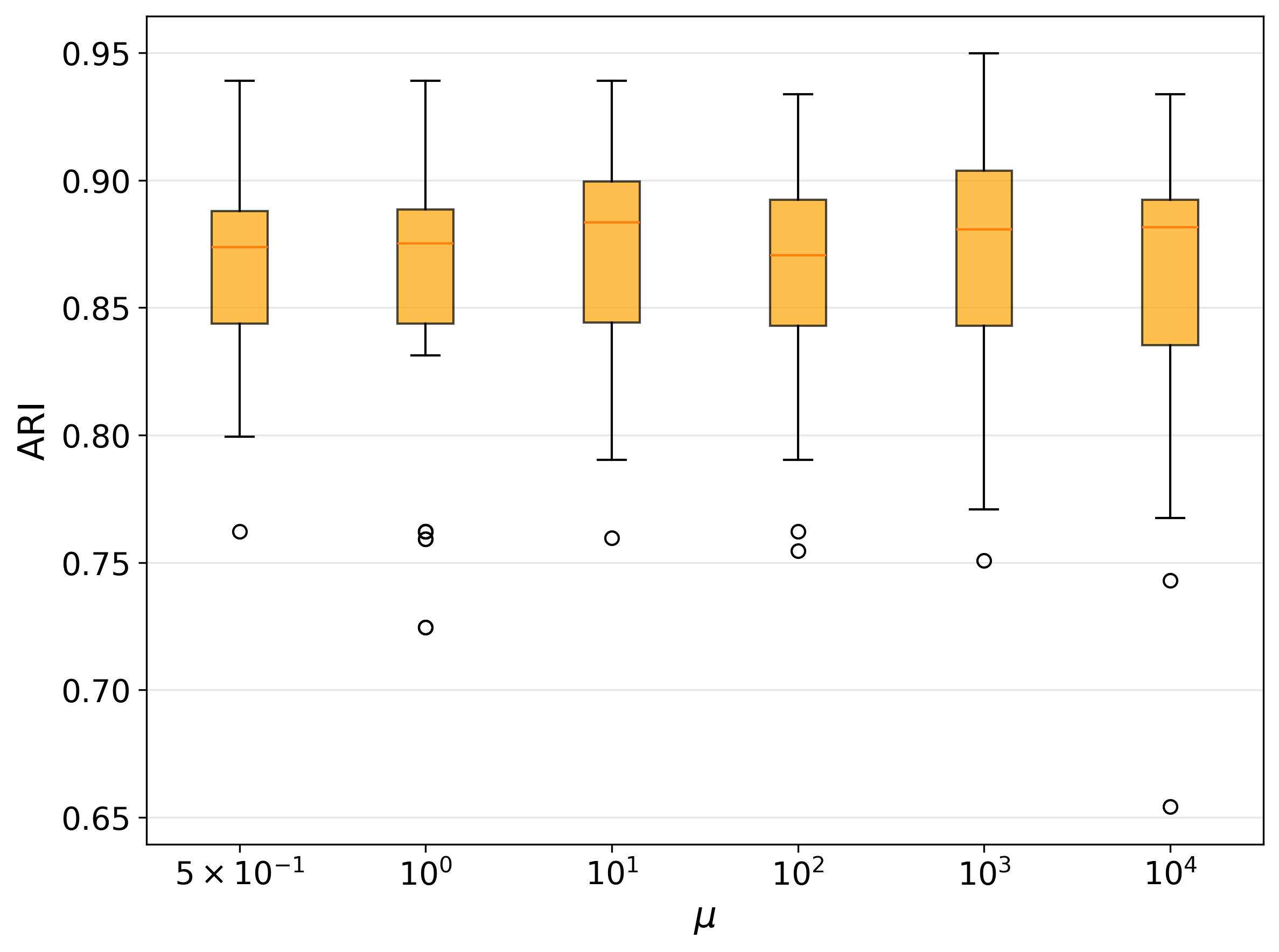}
		\caption{}
		\label{fig:mu_sensitivity}
	\end{subfigure}
	\hfill
	\begin{subfigure}[b]{0.48\textwidth}
		\centering
		\includegraphics[width=\textwidth]{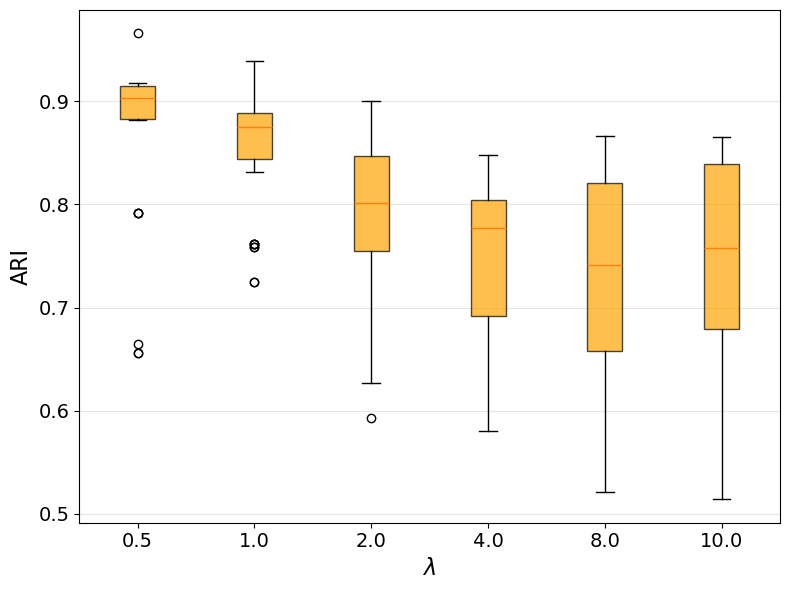}
		\caption{}
		\label{fig:lambda_sensitivity}
	\end{subfigure}
	
	\caption{A box-and-whisker plot of the ARI as the hyperparameters $\mu$ and $\lambda$ are varied. \emph{GNN+REG+CONSTRAINT} was run 5 times (with different seeds) on 10 \emph{small} networks with \emph{medium} density.}
	\label{fig:parameter_sensitivity}
\end{figure}

Within the loss function (equation~\ref{eqn:loss}), the hyperparameters $\mu$ and $\lambda$ adjust the influence of the constraint (equation~\ref{eqn:constraint}) and regularization (equation~\ref{eqn:balance_reg}), respectively. We performed a hyperparameter sensitivity analysis to better understand the influence of $\mu$ and $\lambda$ (figure~\ref{fig:parameter_sensitivity}). Figure~\ref{fig:mu_sensitivity} illustrates a consistent performance as $\mu$ varies, with $\mu=1$ having the lowest variance. The constraint is zero once it is satisfied and, therefore, it is expected that the ARI should not vary much as $\mu$ increases. On the other hand, figure~\ref{fig:lambda_sensitivity} depicts the ARI as $\lambda$ varies and we observe that as it increases, the overall performance decreases (note that the variance increases). This is to be expected since, as $\lambda$ increases, the balance regularization term has more influence during optimization. If its influence is too large, finding communities of equal size would take precedence over modularity and, thus, ARI would decrease. 

By default, we set $\mu=\lambda=1$ for our experiments and reported the results of roughly 2000 synthetic and real experiments, out of which the constraint was violated only once (refer to figure~\ref{fig:varying_lower_bound} and a discussion in section~\ref{subsec:results_syn}). Furthermore, in figure~\ref{fig:cora_varying_lower_bound_1}, the ARI is high with the lowest variance when $\mu=1$. Therefore, we believe that setting $\mu=1$ is a good initial guess for the hyperparameter.

\subsection{Constraint and modularity's resolution limit}\label{subsec:resolution_limit}
A side-effect of our constraint (equation~\ref{eqn:constraint}), coupled with the regularization term (equation~\ref{eqn:balance_reg}), is that it can help counteract the inherent resolution limit of modularity. It is important to note that this is an indirect consequence that can only help in some scenarios. For example, imagine a scenario where there is a small community that is below the resolution limit. Identifying the community would not increase modularity. Now, imagine we know the true number of communities and, as such, set the lower and upper bounds to equal the number of true communities. The constraint would penalize the model until the specified number of communities was found. In other words, even though the small community would result in a sub-optimal modularity score, the overall loss (equation~\ref{eqn:loss}) would be higher because the constraint would be satisfied.

Even though the constraint cannot serve as a general approach to counteract modularity's resolution limit, it may be paired with other approaches that explicitly do, such as $\gamma$-modularity \cite{reichardt2006statistical}.  The $\gamma$-modularity approach uses a hyperparameter $\gamma$ that scales the term $\bold{P}_{ij}$ in equation~\ref{eqn:modularity} and, in so doing, adjusts the size of the community that can be detected. $\gamma$-modularity could replace the standard formulation (equation~\ref{eqn:modularity}) inside the loss function and, thus, serve as a general means to offset the resolution limit.

\subsection{Regularisation terms of MinCutPool and DMoN }\label{subsec:dmon_mincut_reg}

As a reminder, $\bold{S}\in[0,1]^{n \times c}$ is the cluster assignment matrix where $n$ is the number of nodes and $c$ is the number of clusters; $\mathbf{e}_k$ is the canonical vector of $\mathbb{R}^c$ for some $k \in \{1,...,c\}$. Note that $\sum^{c}_{k=1}s_{ik} = 1$ where $s_{ik}\in \bold{S}$

The MinCutPool \cite{bianchi_spectral_2020} penalty term is a normalised version of $||\bold{S}^{T}\bold{S} - \frac{n}{c}\bold{I}||_F$. Note that $||\cdot||_F$ is the Frobenius norm, $\bold{I}$ is the identity matrix, and $\frac{n}{c}$ is the number of nodes per cluster if they were divided equally. See \cite{bianchi_spectral_2020} for the original formulation. The regularization term encourages the row vectors of the cluster assignment matrix to be orthogonal and balanced (the same number of nodes in every cluster). In \citep{tsitsulin_graph_nodate}, the authors show that the MinCutPool regularization term dominates optimization and offer an alternative -- the DMoN penalty term.

The DMoN penalty term is
\[
\frac{\sqrt{c}}{n}\left\| \sum_{i=1}^{n}\mathbf{s}_{i} \right\|_F - 1,
\]
where $\|\cdot\|_F$ is the Frobenius norm and $\mathbf{s}_{i}$ is the $i$-th row vector of $\mathbf{S}$. Similar to the MinCutPool penalty, it was designed to balance clusters. In addition, the authors used it to discourage a trivial solution where all nodes were placed in one cluster. It was proposed as a less dominant penalty term in that the community quality function would be the focus of the optimization, \emph{not} the penalty term. However, it is satisfied (i.e., has its lowest value) both when clusters are balanced and when the cluster assignment is uncertain (i.e., when each cluster has the same probability assignment). We illustrate this mathematically.

In the worst case, when all nodes are assigned to the same cluster $k$, we have
\[
\sum^{n}_{i=1}\bold{s}_{i} = n \mathbf{e}_k 
\] 
and subsequently
\[
\left\| \sum_{i=1}^{n}\mathbf{s}_{i} \right\|_F = n.
\]
Hence,
\begin{align*}
	\frac{\sqrt{c}}{n}\left\| \sum_{i=1}^{n}\mathbf{s}_{i} \right\|_F - 1 &= \frac{\sqrt{c}}{n}\,n - 1\\
	&= \sqrt{c} - 1.
\end{align*}

In the best case (when each cluster has the same number of nodes (i.e., balanced)):
\begin{align*}
	\sum^{n}_{i=0}\bold{s}_{i} = 
	\left[ \begin{array}{c}
		\frac{n}{c} 
		\hdots 
		\frac{n}{c} 
	\end{array} \right] \\
\end{align*}
Hence
\begin{align*}
	||\sum^{n}_{i=0}\bold{s}_{i}||_F &= \sqrt{c (\frac{n}{c})^2}\\
	&= \sqrt{\frac{n^2}{c}}\\
	&= \frac{n}{\sqrt{c}}
\end{align*}

As such
\begin{align*}
	\frac{\sqrt{c}}{n}||\sum^{n}_{i=0}\bold{s}_{i}||_F - 1 &= \frac{\sqrt{c}}{n} \times \frac{n}{\sqrt{c}} - 1\\
	&= 1-1\\
	&= 0
\end{align*}

In the case where the node assignment is uncertain (i.e. $s_{ik}=\frac{1}{c}, \forall i\in\{0,...,n\},k\in\{0,...,c\}$):
\begin{align*}
	\sum^{n}_{i=0}\bold{s}_{i} = 
	\left[ \begin{array}{c}
		\frac{n}{c} 
		\hdots 
		\frac{n}{c} 
	\end{array} \right] \\
\end{align*}
Hence
\begin{align*}
	\frac{\sqrt{c}}{n}||\sum^{n}_{i=0}\bold{s}_{i}||_F - 1 &= 0
\end{align*}
As such, the DMoN penalty term has the same value (i.e., 0) both when the nodes are perfectly balanced across clusters and when the node-cluster assignment is uncertain.

\subsection{Adjusted Rand Index}

To evaluate clustering quality, we report the \textit{Adjusted Rand Index} (ARI), which measures the agreement between a predicted partition and a ground-truth partition while correcting for chance \cite{hubert1985comparing}.
Given two partitions of $n$ nodes, we form a contingency table with entries $n_{ij}$ counting how many nodes are simultaneously assigned to cluster $i$ in the first partition and cluster $j$ in the second.
Let $a_i = \sum_j n_{ij}$ and $b_j = \sum_i n_{ij}$ denote the row and column sums, respectively.
The ARI is defined as
\begin{equation}\label{eqn:ari}
	\mathrm{ARI} = 
	\frac{
		\displaystyle \sum_{i,j} \binom{n_{ij}}{2}
		-
		\frac{
			\displaystyle \sum_i \binom{a_i}{2}\sum_j \binom{b_j}{2}
		}{
			\displaystyle \binom{n}{2}
		}
	}{
		\displaystyle
		\frac{1}{2}\left[
		\sum_i \binom{a_i}{2} + \sum_j \binom{b_j}{2}
		\right]
		-
		\frac{
			\displaystyle \sum_i \binom{a_i}{2}\sum_j \binom{b_j}{2}
		}{
			\displaystyle \binom{n}{2}
		}
	}.
\end{equation}
The ARI takes the value $1$ for identical partitions, is close to $0$ for random unrelated partitions, and can be negative when the agreement is worse than chance.

\subsection{Hyperparameters}

We summarize here the hyperparameters used in our experiments. For the real-world datasets, we employ a 4-layer graph neural network with regularization weights set to $\mu = \lambda = 1$. For the synthetic datasets, we use a 3-layer graph neural network, again with $\mu = \lambda = 1$. In all cases, we use a 2-layer multilayer perceptron. Unless otherwise stated, all models are trained with a learning rate of $10^{-3}$ using the Adam optimizer for 3000 epochs. All experiments were run on Intel Xeon Gold CPUs with 16 cores.

\subsection{Additional synthetic network statistics}
Density is a measure of how close a network is to being fully connected. It is defined as:
\begin{equation}\label{eqn:density}
	\text{Density} = \frac{2m}{n(n-1)}
\end{equation}
where the numerator is the number of edges in the network, and the denominator is the total possible edges in a fully-connected network (with no self-loops).

Moreover, we provide additional statistics in table~\ref{tab:syn_stats_additional} and \ref{tab:cluster_sizes} related to the synthetic networks, namely: the average node degree, community sizes, the $p_{in}$ and $p_{out}$ values used to generate the synthetic networks, as well as the ground-truth modularity. The average node degree is defined as:
\begin{equation}
	\frac{2m}{n}
\end{equation}
The $p_{in}$ and $p_{out}$ from table~\ref{tab:syn_stats_additional} and the cluster sizes from table~\ref{tab:cluster_sizes} can be used to reproduce the synthetic networks using the Python package \textit{PyTorch Geometric} version 2.5.2 \cite{Fey/etal/2025}.

\begin{table}[!htbp]
	\footnotesize
	\centering
	\caption{Additional synthetic network statistics where $p_{in}$ is the intra-cluster edge probability and $p_{out}$ is the between-cluster edge probability. We generated 10 networks for each network type (i.e., there are 10 \emph{small} networks with \emph{low} density). Hence, the average node degree, density, and modularity are approximations.}
	\begin{tabular}{cccccccc}
		\hline
		Size label & Clusters &Density label & Average node degree & Density & $p_{in}$ & $p_{out}$& Modularity\\
		\hline
		Small &  5	& Low			& $\sim$4	& $\sim$0.04& 0.15	& 0.015&$\sim$0.49 \\
		&  		& Medium	& $\sim$11	& $\sim$0.1	& 0.4 		& 0.04&$\sim$0.48 \\
		\hline
		Medium 	&5	& Low 			& $\sim$25	& $\sim$0.02	& 0.1	& 0.002& $\sim$0.67 \\
		&	& Medium	& $\sim$100	& $\sim$0.1	& 0.4 		& 0.007&$\sim$0.67\\
		&	& High			& $\sim$154	& $\sim$0.15		& 0.6		& 0.015&$\sim$0.65\\
		&10	& Medium 	& $\sim$72	& $\sim$0.07	& 0.4 	& 0.002 &$\sim$0.69\\
		&20	& Medium	& $\sim$51	& $\sim$0.05	& 0.8 		& 0.004&$\sim$0.84\\
		
		\hline
		Large & 5	& Low 			& $\sim$103	& $\sim$0.01 & 0.033  & 0.0003&$\sim$0.59 \\
		\hline
	\end{tabular}
	
	\label{tab:syn_stats_additional}
	\vspace{4em}
	
	\footnotesize
	\centering
	\caption{Cluster sizes of each network}
	\begin{tabular}{ccc}
		\hline
		Size label & Clusters & Size of clusters\\
		\hline
		&  		& \\
		Small &  5	& 25, 30, 10, 20, 15 \\
		&  		& \\
		\hline
		&  		& \\
		Medium 	&5	&39, 175, 236, 270, 280\\
		
		&	& \\
		&10	& 10,  23,  27,  32,  38,  97, 108, 170, 229, 266\\
		&  		& \\
		&20	& 20, 31, 71, 73, 29, 19, 21, 32, 15, 65, 60, 53, 76, 80, 70, 27, 62, 61, 85, 50\\
		&  		& \\
		\hline
		&  		& \\
		Large & 5	& 175,  478, 2358, 2989, 4000 \\
		&  		& \\
		\hline
	\end{tabular}
	
	\label{tab:cluster_sizes}
\end{table}

\begin{table}
\end{table}
\begin{table}
\end{table}
\begin{table}
\end{table}
\begin{table}
\end{table}
\begin{table}
\end{table}
\begin{table}
\end{table}

\end{document}